\documentclass[manuscript,screen]{acmart}

\AtBeginDocument{%
	}

\setcopyright{acmlicensed}
\copyrightyear{2018}
\acmYear{2018}
\acmDOI{XXXXXXX.XXXXXXX}

\acmConference[Conference acronym 'XX]{Make sure to enter the correct
	conference title from your rights confirmation emai}{June 03--05,
	2018}{Woodstock, NY}
\acmISBN{978-1-4503-XXXX-X/18/06}

\acmSubmissionID{123-A56-BU3}

\begin{document}
	
	\title{OneTo3D: One Image to Re-editable Dynamic 3D Model and Video Generation}
	
	\author{Jinwei Lin}
	\email{Jinwei.Lin@monash.edu}
	\orcid{0000-0003-0558-6699}
	\authornotemark[1]
	\affiliation{%
		\institution{Monash University}
		\city{Clayton}
		\state{Victoria}
		\country{Australia}
	}
	
	

	\renewcommand{\shortauthors}{Jinwei et al.}
	
	\begin{abstract}
		One image to editable dynamic 3D model and video generation is novel  direction and change in the research area of single image to 3D representation or 3D reconstruction of image. Gaussian Splatting has demonstrated its advantages in implicit 3D reconstruction, compared with the original Neural Radiance Fields. As the rapid development of technologies and principles, people tried to used the Stable Diffusion models to generate targeted models with text instructions. However, using the normal implicit machine learning methods is hard to gain the precise motions and actions control, further more, it is difficult to generate a long content and semantic continuous 3D video. To address this issue, we propose the OneTo3D, a method and theory to used one single image to generate the editable 3D model and generate the targeted semantic continuous time-unlimited 3D video. We used a normal basic Gaussian Splatting model to generate the 3D model from a single image, which requires less volume of video memory and computer calculation ability. Subsequently, we designed an automatic generation and self-adaptive binding mechanism for the object armature. Combined with the re-editable motions and actions analyzing and controlling algorithm we proposed, we can achieve a better performance than the SOTA projects in the area of building the 3D model precise motions and actions control, and generating a stable semantic continuous time-unlimited 3D video with the input text instructions. Here we will analyze the detailed implementation methods and theories analyses. Relative comparisons and conclusions will be presented. The project code is open source.  
		
	\end{abstract}

	
	
	
	\maketitle
	
	\section{Introduction}
	
	3D representation or 3D reconstruction is a challenging for a long time in the research area of computer vision. The current methods to achieve the 3D reconstruction can be divided into two main classes, one is using the traditional methods to directly design and complete the 3D reconstruction or modelling, which are usually classified as explicit methods; the other is using the machine learning methods and theories to achieve these goals, which are classified as implicit methods. During the development of these years, one outstanding and novel technological revolution is the Neural Radiance Fields (NeRF) \cite{mildenhall2021nerf}, in which the researchers used the special designed 5D coordinates representation function with classic volume differentiable rendering techniques and the machine learning of Multi-layer Perception (MLP). NeRF can obtain a better performance in the rendering and representation of photo-realistic scene views. Based on NeRF \cite{mildenhall2021nerf}, various of research projects about implicit 3D representation or reconstruction has developed booming.
	
	In the early stages of development of NeRF-based researches, most of the research project focused on implementing a better static NeRF (e.g.: Mip-nerf \cite{barron2021mip}, researching on anti-aliasing performance of NeRF rendering; Mip-nerf 360 \cite{barron2022mip} and Nerf in the wild \cite{martin2021nerf}, researching on improving the performance of NeRF in unconstrained or unbounded scenes). Similar to the development and evolution pattern of traditional 3D modelling and reconstruction technologies, researchers gradually been interested in dynamic and editable NeRF (e.g.:D-nerf\cite{pumarola2021d}, adding the time as additional input and change domain to be dynamic; Nerf-editing\cite{yuan2022nerf}, using the explicit methods to edit the implicit representation of the meshes). As the developing of NeRF, another outstanding and novel technological revolution has being happening, using the 3D Gaussian Splatting \cite{kerbl20233d} with machine learning to implement novel and better performance of implicit rendering of 3D representation and reconstruction. Similar to the development and evolution pattern of NeRF, people started to study the implementations of dynamic and editable 3D Gaussian Splatting (e.g.: 4d gaussian splatting \cite{wu20234d}, using the HexPlane to adding the time domain and building 4D neural voxels). Compared with NeRF, 3D Gaussian Splatting is advantaged in high resolution, photo-realistic rendering and rapid processing speed. As the developing of Gaussian Splatting, another robust technological revolution has being happening, using the Stable Diffusion \cite{rombach2022high} models to achieve text to 3D or text to 3D video, which can gain higher resolution and more powerful and creative machine design ability but requires larger computing resource and time. Using the Stable Diffusion model to generate 3D model or video is mainly a pure implicit method. For all of the technologies mentioned above, or the existed similar research project, it is still difficult to achieve the goal of implementing a precise and detailed specific motion and actions editable and controlling of the dynamic and re-editable 3D model or video with a high processing performance.
	
	To achieve a robust method for the dynamic and re-editable 3D reconstruction, 3D representation, 3D generation for the 3D models and videos, we combined the novel implicit machine learning rendering or reconstruction method, and the traditional explicit modelling and controlling idea, proposed the OneTo3D. We selected the Zero-1-to-3 \cite{liu2023zero} as our basic 3D generation model, to build a initial basic 3D model. This model is weak in generation reduction degree and resolution but requires less computing VRAM and processing ability, compared with Sv3d\cite{voleti2024sv3d} that used Stable Diffusion. In the first stage of generating the inital 3D model, OneTo3D is model-independent, which means other similar model can be also used in OneTo3D. The first version of OneTo3D is focusing on make the generation of 3D model be re-editable and dynamic, meanwhile gain the targeted 3D video generation. 
	
	After gaining the initial 3D model, the next operation is automatically generating the self-adaption armature for the object. To easier verify the theory and implement the experiments, we selected the human type objects as the test object. The armature automatic generation algorithm will analyze the most suitable armature pose for the tested object and generate the self-adaption armature automatically, following by the automatic bones binding between the object and armature. Cooperated with motion and actions analyzing algorithms, the user input text instructions will be analyzed and translated to detailed controlling command list that will be applied in the edition and controlling the motions and actions of the object. Combining with Blender application and the key frames deduction algorithms, generating the final 3D model or 3D video. Compared with other pure implicit or pure explicit 3D representation or reconstruction research projects, OneTo3D has combined the advantages of the rapid rendering speed and high resolution of implicit 3D rendering and edition, and the advantages of re-editable and easy and precise dynamic controlling of explicit 3D rendering and edition. Compared with multi-views input images to 3D generation, one image to 3D generate has been more challenging and user-friendly. Compared with other SOTA research projects in the area of implicit 3D representation and reconstruction, OneTo3D gained better performance in dynamic rendering speed and more precised controlling fineness in re-editable operations. More analyses for the experiments and theories will be presented.

	\section{Background}
	
	The developing and evaluation of implicit 3D modelling and generation is fast, as the main representatives, 3D model and 3D video generation is one of the most challenging, which is usually represented by 3D representation and 3D reconstruction. Our research focus on the editable and dynamic 3D generation. 
	
	\subsection{Explicit and Implicit}
	
	In the current research area of 3D representation and reconstruction, whether using a machine learning method to implement the 3D rendering and representation, is a general criteria to evaluate a 3D rendering and representation method is implicit or not. Compared with the implicit 3D generation or representation, explicit 3D generation or representation has a longer history of development. The traditional 3D generation technologies are usually classified as the explicit ones, which mainly directly process the 3D data with the calculation of mathematical and physical properties of the 3D models, likes meshes \cite{zhang2001efficient}, cloud points \cite{kazhdan2006poisson}, photos or images \cite{snavely2006photo}, radial basis functions \cite{carr2001reconstruction} and camera \cite{zollhofer2018state}. The implicit 3D generation or representation projects are usually designed combing with machine learning. Although NeRF has gained a great performance and impact in 3D generation or representation area, it is not the first research project pioneer of implicit 3D generation or representation. Using the learning network to generate a 3d reconstruction with one image is achieved before \cite{fan2017point}. For implementation methods, Mvsnet \cite{yao2018mvsnet} and Occupancy networks \cite{mescheder2019occupancy} are two of the early classic representatives of using the deep learning. For cloud points, Pointrcnn \cite{shi2019pointrcnn} is is one of the early classic representatives of using the learning networks. People used the methods of machine learning to gain more detailed, complex or exquisite representations of 3D generations. As the developing of deep learning technologies in 3D generation or representation, more research groups focus on using the pure learning methods to implement the 3D generation or representation (e.g.: NeRF, Gaussian Splatting and Sv3D, etc.). 
	
	However, using the pure learning network methods is difficult to achieve a great performance in specific detailed and precise controlling in edition and dynamic motion. Purely using the machine learning to make the 3D generation is seeming more advanced but have not gained the better performances in many areas of 3D generation or representation, for example, re-editable 3D model edition, detailed and precise motion or actions controlling of the 3D models, larger requirement in computing resource, and longer processing or rendering time. Throw out the traditional 3D reconstruction or representation technologies completely and replacing them with learning networks is not unwise. Some of the traditional 3D generation or edition technologies are mature, reliable, and not necessary be replace with learning network. A current suitable and better solution is combing the advantaged of implicit and explicit methods of 3D generation and representation, which is also adopted in this version of OneTo3D. We will make use of explicit methods, relative machine learning methods, combining with specific novel methods, to realize re-editable character of OneTo3D.

	\subsection{Re-editable and Dynamic}
	
	Whether can be editable is usually one indispensable evaluation criterion for great 3D modelling and generation. An editable character means the generated 3D model can be edited. We classified this editable character into two types: one-off editable character or onece-editable character, and the re-editable. One-off editable character means the edition process can not be suspended or multiple modified during the processing, the unfinished results from mid-edit can not be saved for the next edit; re-editable character represents the edition process can be multiple or continuously suspended or modified during the processing, the unfinished results from mid-edit can be saved for the next edit. Obviously, it will be more hard obtaining a re-editable character rather than the one-off editable character. The editable character of OneTo3D belongs to re-editable. Almost all of the existed NeRF or 3D Gaussian Splatting or Stable Diffusion research projects we surveyed are one-off editable. For example, for NeRF: D-nerf \cite{pumarola2021d}, can control the scene camera view and motion happen time only following the fixed dynamic sequence; this type of synthesizing dynamic models \cite{gao2021dynamic} is also only following the existed dynamic monocular video sequence; Hdr-nerf \cite{huang2022hdr}, controlling the different exposures to control the ray origin or direction of the rendering scene. For 3D Gaussian Splatting: 4D gaussian splatting \cite{wu20234d}, using a neural HexPlane-based voxel encoding algorithm to predict the deformations, the rendering contents can not be re-editable when the scenes are completely rendered. Purely using implicit methods will require more calculation resources and spend more processing time. It needs more developing time for current implicit methods to achieve a real-time processing ability in 3D re-editable edition. Even depending on the robust Stable Diffusion models, Sv3D \cite{voleti2024sv3d}, is still unable to achieve a pragmatic near real-time re-editable processing. However, there are still few projects that have implemented the re-editable of 3D by using the Gaussian Splatting, which have combined the other various machine learning technologies and traditional computer graphic processing methods, e.g.:Gaussianeditor \cite{chen2023gaussianeditor}, using the Hierarchical Gaussian Splatting semantic tracing technology to achieve the re-editable processing. 
	
	Nevertheless, the re-editable processing ability of Gaussianeditor is mainly focus on special application scenes, and unable to make the edition of dynamic scenes. Due the strong generalization ability, generating confrontation ability and independent innovation design ability of Stable Diffusion models and Transformer models, various of text-to-3D research project are presented, e.g.: Zero-1-to-3 \cite{liu2023zero}, which is based on threestudio \cite{threestudio2023}, which is developed based on a large-scale conditional diffusion model and a 3D synthetic dataset. We tested the Zero-1-to-3 and found it is able to generate most of the basic details of targeted model, but still weak in generating a precise model. Using the Sv3D \cite{voleti2024sv3d} will gain better reconstruction in precise details but need more VRAM and processing time. Extending the volume of the training dataset will improve the detailed reconstructing and representing ability of the large model, e.g.: Objaverse-XL \cite{objaverseXL}, which is developed based on a large vloume 3D dataset of over 10 million 3D objects and Zero-1-to-3 \cite{liu2023zero}, and gained better performance in generating 3D models. Our research of this version of OneTo3D will focus on combining the implicit 3D modelling, traditional re-edition relative technologies and novel text-instructions to 3D motion and actions analyzing and implementing mechanism, to achieve the near real-time re-editable functions.

	\subsection{Motions and Actions}
	
	Motions and Actions are two main deformation characters of 3D model or video generation. In the traditional explicit 3D representation or reconstruction, deforming a 3D object is usually changing the shape of the object. One main methods are used to changing the shape of 3D object: changing the representation and location of the 3D unit, e.g.: meshes or point clouds. Other influence factors should be also considered, e.g.: environment lights, reflection of and cameras. Purely using the implicit methods to control the motion or action of the generated 3D model is difficult and requires large computing resources, compared with the explicit methods, because of the implicit methods need to render or reconstruct many continuous frames one by one, for example the Sv3D \cite{voleti2024sv3d} or 4D gaussian splatting \cite{wu20234d}. Another feasible solutio is using the machine learning method of Stable Diffusion \cite{rombach2022high} or other Generative Adversarial Network (GAN) \cite{goodfellow2014generative} relative models to make the targeted controlling and edition for the 2D image of the final 3D model first, which needs stable diffusion models or generation model has robust semantic analyzing and image content processing abilities. Meanwhile, the continuity and consistency of the image content should be remained. Subsequently, using the 3D generation model to generate the targeted 3D model with specific action frame by frame, that is why the models like Sv3D \cite{voleti2024sv3d} or 4D gaussian splatting \cite{wu20234d} need long time to generate several seconds 3D video.            
	
	In OneTo3D, we use the traditional explicit methods to implement the re-editable controlling and edition of the 3D model pose to control and edit the motions and actions of the targeted model. If using the explicit methods to control the preside and detailed 3D graphic industrial grade deformation of the 3D model, a recommendable method is developing based on mature 3D editable framework or toolkit that is open source. We selected the Blender \cite{hess2013blender} as the basic indirect 3D edition framework, cooperated with novel armature self-adaption generation algorithm and text-to-actions analyzing algorithm, to make the re-editable 3D generation. We used the key-points detection toolkit of MMPOSE \cite{mmpose2020} to gain the key-points information for our designed algorithm of armature automatic generation and self-adaption. Our idea is generating a suitable armature for the 3D model first, and using the text instructions to control the poses of the armature, to directly control the 3D model making specific motions and actions. Controlling the armature \cite{barber2003survey} with Blender, can achieve precise and detailed controlling of the motions and actions of the 3D model in any tiny level supported by the 3D framework. Another research we selected Blender is we can make direct interaction between Blender and our algorithms written with Python easily \cite{conlan2017blender}. Various of traditional tool-kits or auxiliary algorithms can be used in processing, like changing format of 3D model , generating the 3D video and re-editable Blender files \cite{hess2013blender}.

	\section{Methodology}
	
	Here, we will make the detailed analyses of the design and implementation of OnTo3D. The core are self-adaption 3D model armature generation, the instruction semantic extraction and the implementation of text-to--action poses.

	\subsection{Architecture}
	
	\begin{figure}[t]
		\centering
		\includegraphics[width=0.9\columnwidth]{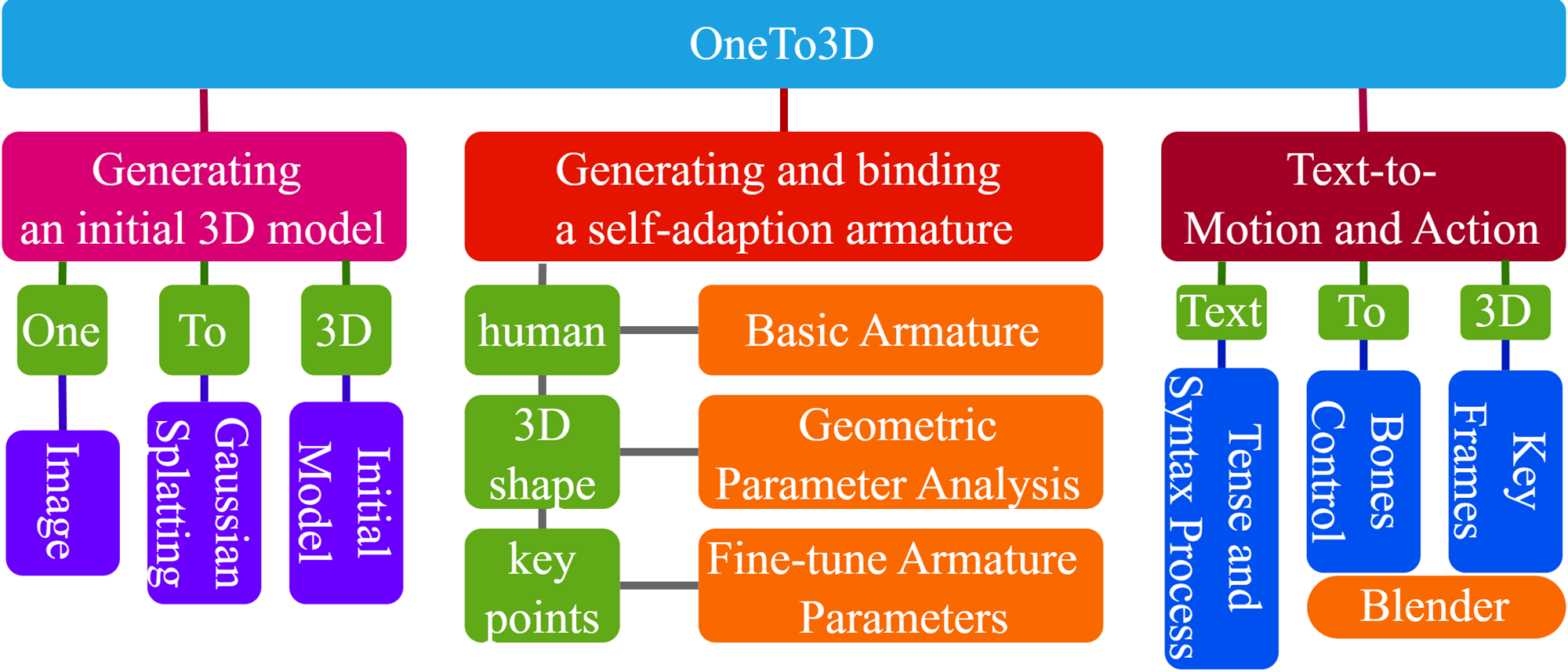}
		\caption{Architecture of OneTo3D.}
		\label{fig1}
	\end{figure}

	As shown in Figure \ref{fig1}, the main architecture of OneTo3D can be divided into three parts: Generating an initial 3D model, Generating and binding a self-adaption armature, Text-to-Motion and Action, which also corresponds to the three main sequential processing phases. The aim of phase generating an initial 3D model, is to gain the basic and initial 3D model of the input one image. There is no dynamic or editable factor in the initial 3D model. The initial input is one image that contains the whole body pose and shape descriptions of the model, which usually is a front view of the object. Then the image will be handled with a series of pre-processing, including the main semantic content detecting and capturing, removing the background and side-views analyzing. Subsequently, the Gaussian Splatting model will handle the processed image and generate a initial 3D model. The initial 3D model generation part of OneTo3D is based on DreamGaussian\cite{tang2023dreamgaussian}, which relies on Zero-1-to-3 \cite{liu2023zero}. 
	
	The aim of phase generating and binding a self-adaption armature, is building a suitable human body type armature to control the generate initial object body with Blender. First, we designed a basic novel armature that fits the normal human body. Following, we analyzed the geometrical parameters information of the shape of initial 3D model, including the width, height, as well as the rotations, locations, and lengths of each bones of the generated armature. 
	In this process, the initial 3D model will be input into the Blender, and the basic armature will be built with the 3D model in same work space in Blender. Subsequently, OneTo3D will analyze and calculate the best data to fine-tune the initial model, with the pose, shape, and key points information that are gained from the input single image. Fine-tuning the armature parameters until the generated armature is fit the body of the object.   
	
	The aim of phase text-to-motion and actions, is analyzing the command intention of the user text instruction, and translate the commands specific motion and modification data of relative bones of armature. We design an special analyzing algorithm to analyze the tense and syntax of the user input commands, then extracting the locations and direction degree data of the relative bones, following by controlling the specific bones to implement relative motion in Blender. We add parameters consideration and analyzing of motion quantification, the number of motion times, the motion direction and the movement range, etc. The commands interpreter mechanism supports multiple sub-command splitting and continuous commands edition. Multi-commands will be split into independent sub-command, and each sub-command can be translated into one specific control information that contains the detailed motion data. 
	
	If there is no specific motion range or quantization data contained or provided in the user command, the command will be translated with default motion data, e.g.: the rotation degrees, motion times, or motion directions generated randomly. The motion re-editable controlling is implemented cooperated with Blender interface. After controlling the armature to move following the targeted pose, using the coding to insert the current pose as a key frame. Combining the continuous key frames to generate the final 3D video. The Blender files will be saved as re-editable 3D edition file.

	\subsection{Generation}
	
	\begin{figure}[t]
		\centering
		\includegraphics[width=0.9\columnwidth]{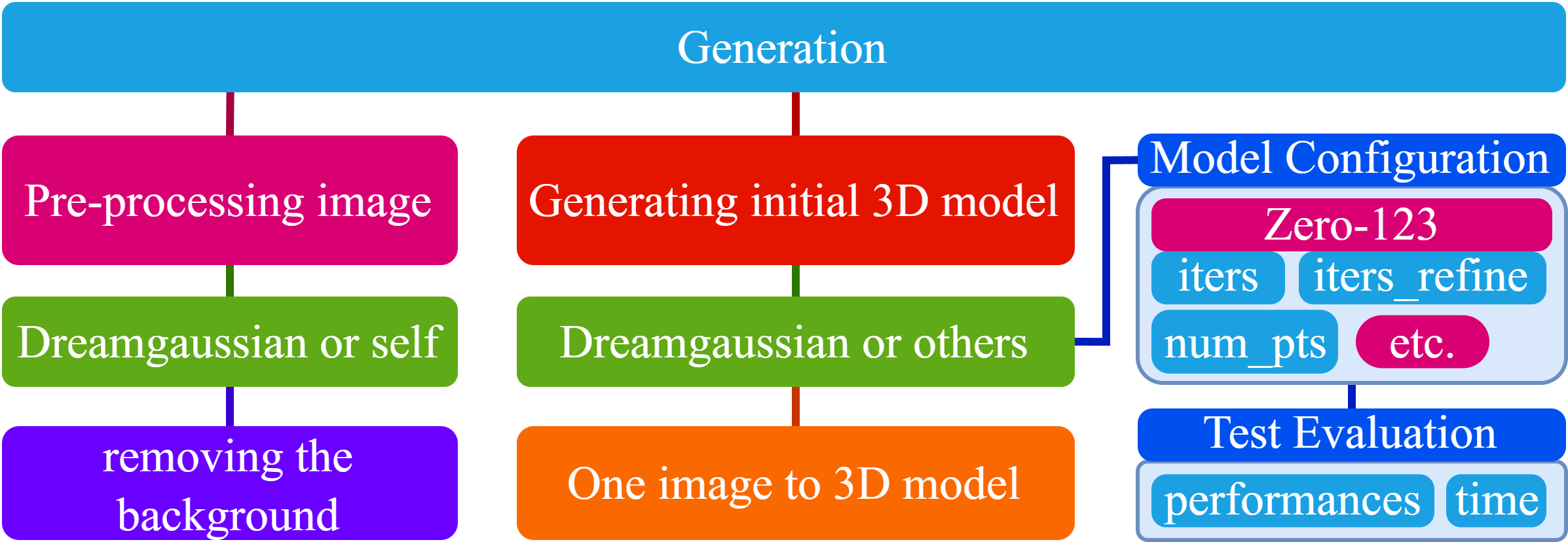}
		\caption{Architecture of OneTo3D.}
		\label{fig2}
	\end{figure}
	
\begin{figure}[h]
		\centering
		
		\begin{minipage}{0.33\linewidth}
			\centering
			\includegraphics[width=0.9\linewidth]{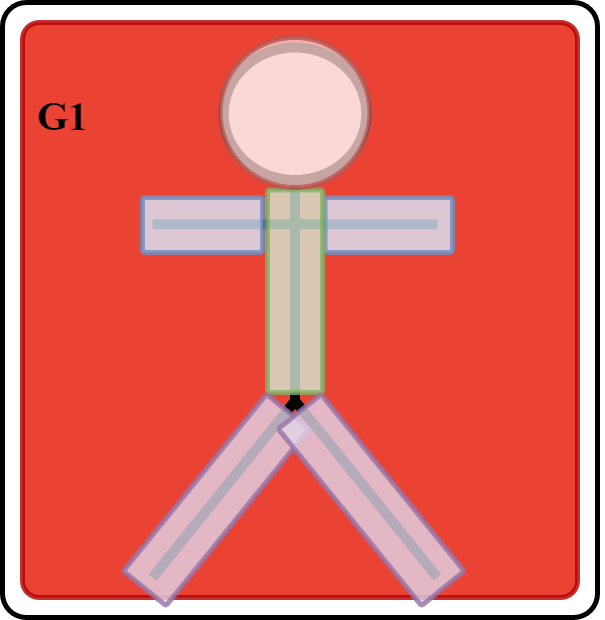}
			\centerline{(a) image with one pure color group.}
		\end{minipage}
		\begin{minipage}{0.33\linewidth}
			\centering
			\includegraphics[width=0.9\linewidth]{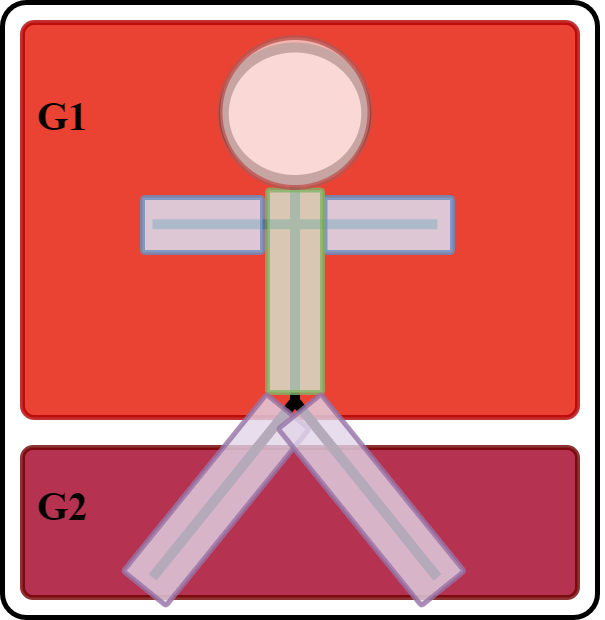}
			\centerline{(b) image with 2 color groups.}
		\end{minipage}
		\begin{minipage}{0.33\linewidth}
			\centering
			\includegraphics[width=0.9\linewidth]{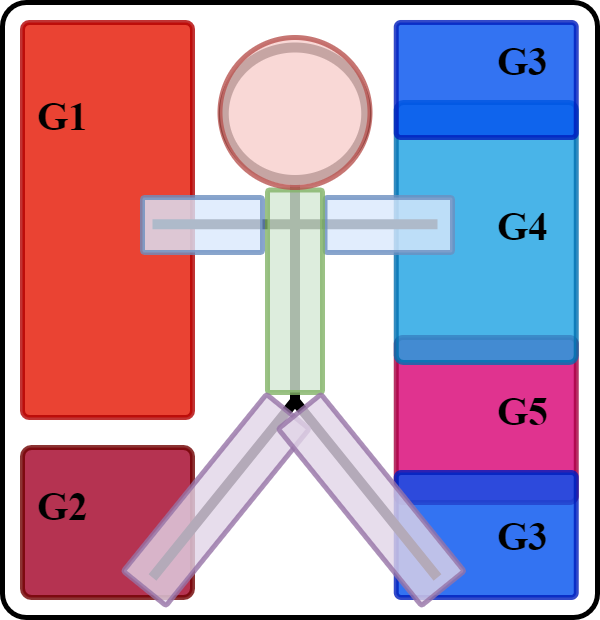}
			\centerline{(c) image with multiple color groups.}
		\end{minipage}
		
		\caption{Removing the background with color groups.}
		\label{fig3}
	\end{figure}

	The first main stage of the processing of OneTo3D is generating the initial 3D model.  For the input image, it is optional to use text-to-image stable diffusion relative generation models to generate a targeted object image. As shown in Figure \ref{fig2}. For removing the background, it is available to choose the $process.py$ script of Dreamgaussian or using others. Removing the background will reducing the disturbance from the effect-irrelevant image noises. There are various methods or algorithms that can remove the background of the images, or using the image detection or semantic segmentation machine learning methods. Here, we provides two optional methods for removing the background. The first method is designed for the simple background image, it is available to calculating the proportion of each existed main color items in the image, and dividing the color items that are in a color values range into different color groups, removing the color groups that are in a configured proportion range. As shown in Figure \ref{fig3}, a recommended input object image is an object image with a pure color likes $image1$. For other images that have multiple color blocks in the input image, like $image2$ and $image3$, etc., this method is also available. Parameters $G_{i}, i \in \mathbb{N}^+$ represent the color groups.

	As shown in Equation \ref{eq1}, using the $(r_{i}, g_{i}, b_{i}$ to present the RGB format color values of the pixel with index $i$. In sub-equation $model sort$, if the pixel color value is the same with the existed one, count it adding 1 and record as $[(r_{i}, g_{i}, b_{i}), count(i)+1]$. Make the iterative calculations for all of the pixels of the input 2D image, following by sort them according to the index $1$ item $count(i)$, to gain the sorted color value list. There are two main operating equations to describe the processing of dividing the groups. In sub-equation $divide\ groups1$, select the $ first N largest numbers$ that have the first $n$ largest numbers of $count(i)$ as the segment groups and represent them as $sg$. 
	
	\begin{equation}
		\label{eq1}
		\left\{
		\begin{aligned}
			mode\ sort :\rightarrow sort\_colors = sort(\sum_{i}^{N}{if(r_{i}, g_{i}, b_{i})\Rightarrow [ first N largest numbers, count(i)+1], i \in \mathbb{N}^+})   \\
			divide\ groups 1: \rightarrow sg:sg_{i} = (r_{i}, g_{i}, b_{i})\ | \  if(i <= n)     \\
			divide\ groups 2: \rightarrow jg:jg_{i} =  \left\{ \begin{aligned}
				judge(if(r_{i}, g_{i}, b_{i}) \in (r_{i} \pm R_{r}, g_{i} \pm R_{g}, b_{i} \pm R_{b})) \Rightarrow merge  \\
				\ | \  R_{j,j\in (r, g, b)} \in \mathbb{N}^+);group_{i} (r_{i}, g_{i}, b_{i}) \in sg  \\
			\end{aligned} 
			\right.   \\
			remove\ groups: \rightarrow remove\_sort\_colors = (group_{i}, i <= n, i \in  \mathbb{N}^+) | group_{i} \in jg   \\
		\end{aligned}
		\right.
	\end{equation}
	
	In sub-equation $divide\ groups2$, judging the color values $(r_{i}, g_{i}, b_{i})$ for each pixel, if the RGB color component values satisfy the positive and negative floating range $\pm R_{r}, \pm R_{g}, \pm R_{b}$, than merging the pixels into a color group. In sub-equation $remove\ group$, after gaining the color groups list $jg$ that are processed by judging and merging, directly removing the pixels in the first $n$ groups of $jg$, the result image is the targeted. If the background-removing result is still not satisfied for the requirement, changing the value of parameter $n$ to get better processing result. This method is designed based on judging the available quantization ratio of the color components of the image to remove the background, which is propitious to remove the background color component that has simple, monotonous or monochromatic background. 
	
	For removing the background, the second method we provided is designing the background removing algorithm with some suitable machine learning method idea, e.g.: edge detection and armature or body key-points detection, etc., which are beneficial for differentiating the main image area more efficiently.

	\begin{figure}[h]
		\centering
		
		\begin{minipage}{0.33\linewidth}
			\centering
			\includegraphics[width=0.9\linewidth]{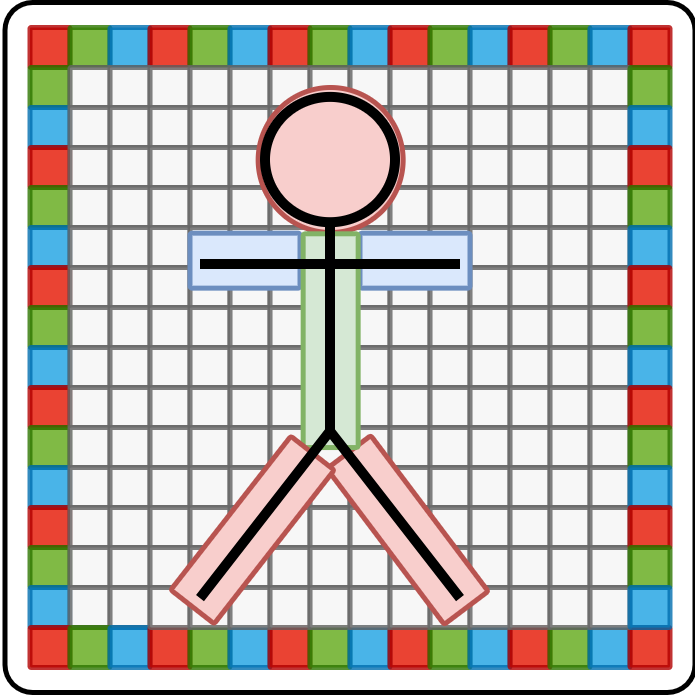}
			\centerline{(a) object with complex background.}
		\end{minipage}
		\begin{minipage}{0.33\linewidth}
			\centering
			\includegraphics[width=0.9\linewidth]{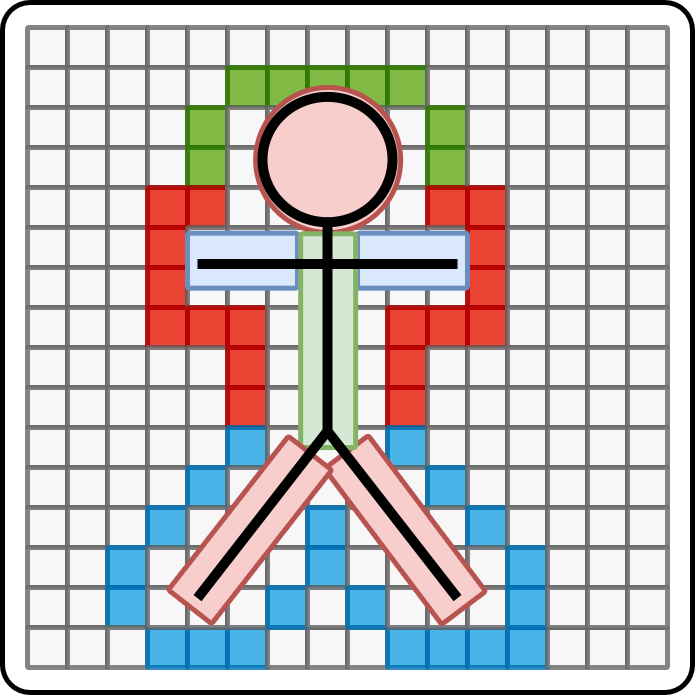}
			\centerline{(b) removing with edges detection.}
		\end{minipage}
		\begin{minipage}{0.33\linewidth}
			\centering
			\includegraphics[width=0.9\linewidth]{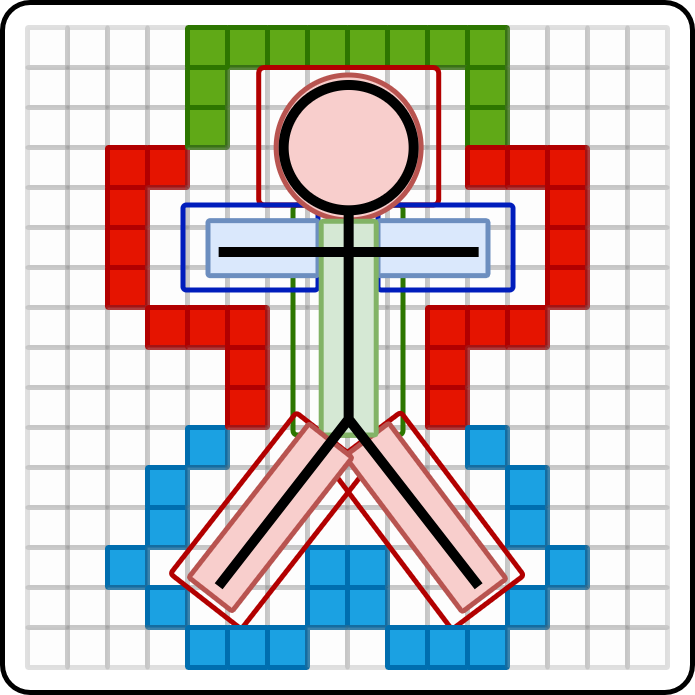}
			\centerline{(c) removing with trunk domain.}
		\end{minipage}
		
		\caption{Removing the background with edge detection and armature structure or body key-points detection and trunks domain segmentation. The background is usually colorful and complex in RGB other formats. Using a block with one single color to present a colorful pixel. Black bold simple human-like outline present the designed novel armature of the object, which including the key points detection data. Sub-figure (c) represents the trunks domain that contains the main components of the object trunk.}
		\label{fig4}
	\end{figure}
	
	As shown in Figure\ref{fig4}, when removing background of the input object image that has a complex background, purely using the method one will not satisfy the requirement. As shown in sub-figure\ref{fig4}-a, although the background of the image maybe complex and colorful, for those images that has different color between the trunk parts areas and the background behind, it is available to remove the background. For this issue, we presents two methods: one is removing the background with edge detection, i.e.: detecting the whole continuous outline edge of the object, following by removing the color pixels that locate outside the object outline edge area; the other method is using the machine learning methods to detect the key points of the object first, subsequently, defining a trunk domain area for each main part of the body armature, e.g.: arms, legs and head, etc. The sizes of the trunk domain have default values and are available to be customized. The principle of defining the trunk domain is insuring all of truck parts are enclosed inside the trunk domains. Removing the trunk domains outside color pixels after the trunk domains division.       
	
\begin{figure}[t]
		\centering
		\includegraphics[width=0.99\columnwidth]{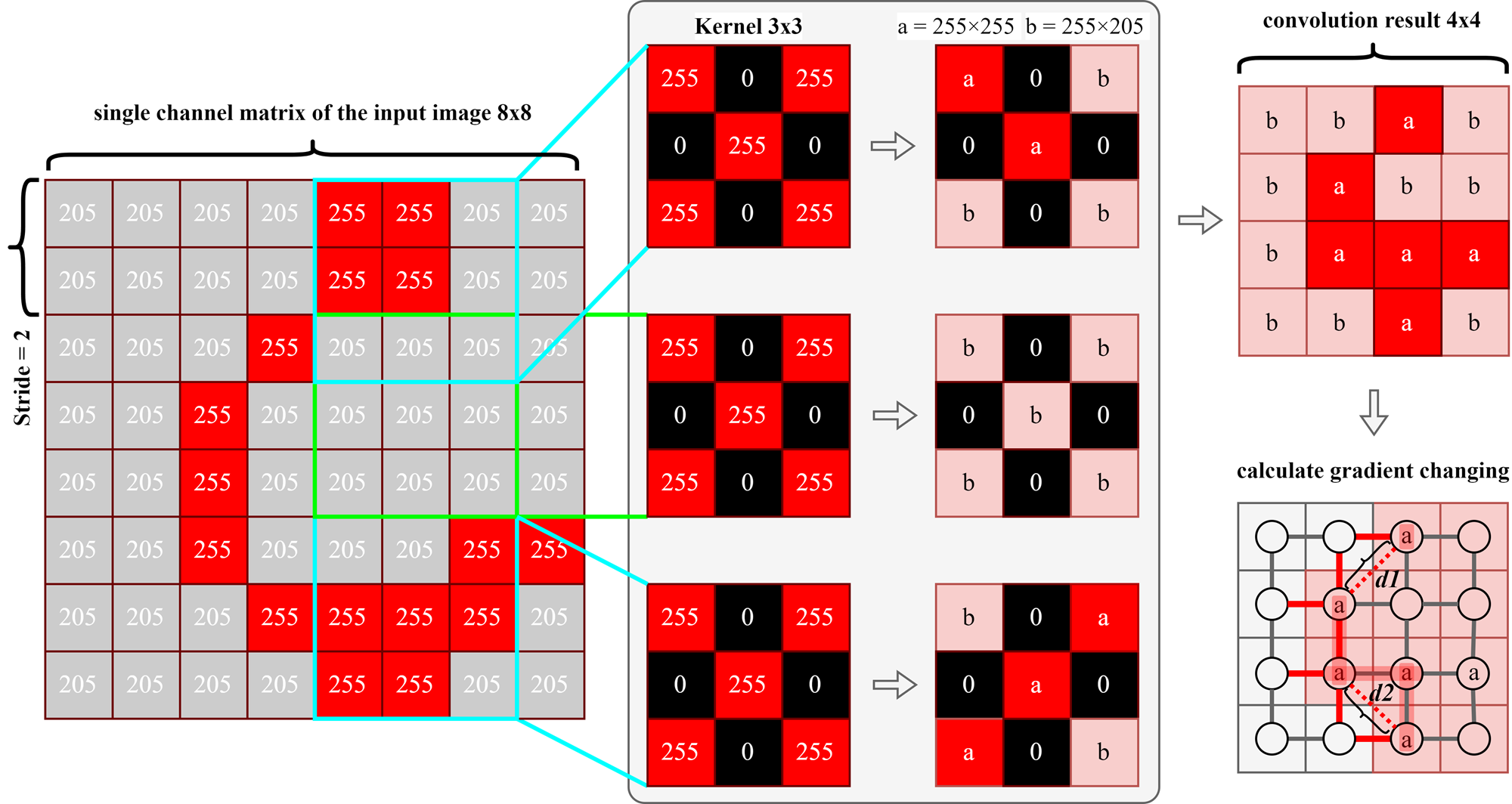}
		\caption{Removing the background with edges detection based on convolution computation gradient changing calculation.}
		\label{fig5}
	\end{figure}

	The implementation principle of removing the background with edges detection can be described as shown as Figure \ref{fig5}, which takes the processing of single red color channel as example. First, using the convolution computation to extract and compress edge features information. $stride$ represents the stride of the convolution computation, which is customized. We designed a convolution kernel with size $3\times 3$ to extract the color features of the quadrangle and center, where have the default customized value 255, the values elsewhere are 0. After the convolution computation, the compressed edge features information is gained as the convolution result. Between the two sides of the edge of the object, the gradient changing level is different, therefore we designed based on this to judge the edge condition.

	The above implementation principle can be described further detailed with Equation \ref{eq2}, $R(i, j)$ represents the result of the convolution computation. Parameters $w$ and $h$ represent the width and high dimensions of convolution core matrix $K_{\alpha, \beta}$. Usually $w=h$. Matrix $I$ is a square matrix usually with a dimension $m$. The dimension of the result matrix is $n$. Parameter $p$ represents the padding number of the edge of $I$. Using the specific core matrix $K_{\alpha, \beta}$ to make the matrix multiplication with $I$, and name the result as $K_{i, j}$. Subsequently, for each multiplication result $K_{i, j}$, judging the result to see if it is larger than $\delta$, recorded the times when the result value is larger than $\delta$. For each unit of the convolution result matrix, if the multiplication result of the unit is not less than $\delta$, name the unit as $a$ unit, otherwise name as $b$ unit. All of the $a$ units consists the set $A$, all of the $b$ units consists the set $B$. To better simplifying and extracting the edge information, we defined two customized threshold values. Parameter $\delta$ is the first threshold value that is used to divide the multiplication result matrix sets $A$ and $B$. Following, enter the stage of calculating the gradient changing.

	\begin{equation}
		\label{eq2}
		\left\{
		\begin{aligned}
			R(i, j) = \sum_{\alpha = 0}^{w} \sum_{\beta = 0}^{h} K_{i,j} = K(\alpha, \beta) * I(i-\alpha, j-\beta) \ | \  \alpha=\beta || \alpha+\beta=w; n = \frac{m-w+2p}{s}+ 1  \\
			R(i, j) \in A\ | [\ if(T>\varphi ) \ | \ T=\sum_{\alpha = 0}^{w} \sum_{\beta = 0}^{h} p \ | \ if(K_{i,j} >= \delta):q =1, else:q=0;]\ else: R(i, j) \in B   \\
			EA = set(a_\sigma) | \sum_{\sigma=0}^{len(A)}\sum_{d=0}^{3} if(G(a_\sigma, d) > \eta) \to G(a_\sigma, d) = 1; \ else: G(a_\sigma, d) = 0   \\
			w, h, m, n, a, b, \alpha, \beta, i, j, T, \varphi, p, q, \delta, \sigma, d, \eta \in \mathbb{N}; a \in A, b \in B   \\
		\end{aligned}
		\right.
	\end{equation}

	Using $\sum_{\sigma=0}^{len(A)}$ to all of the units in matrix $A$, which are the possible edge pixels. For each unit pixel in matrix $A$, defining four possible edge extending directions, i.e.: up, down, left and right. Using the $\sum_{d=0}^{3}$ to count the number how many possible has a large enough gradient changing range. Function $G(a_\sigma, d)$ calculates the gradient changing range between the unit pixel $a_\sigma$ with index $\sigma$ and one of the four sides unit pixel with the direction $d$. he second threshold value is parameter $\eta$ that is used to whether judge the gradient changing range between the unit pixel $a_\sigma$ and the unit pixel that locates in the direction $d$ of $a_\sigma$, is enough large to be considering the unit pixel $a_\sigma$ as one pixel of the edge. Using the set $EA$ to represents the collection of all of the analyzed edge points. In the sub-graph of calculating the gradient changing of Figure \ref{fig5}, using the red bolder lines to represent these gradient changing level between two adjacent points, is large enough to record, while using the gray bold lines to represent the gradient changing level between the gradient changing level between two adjacent points, is large enough to record, two adjacent points, is not large enough to record. To improve the calculation effect and edge detection details, one method is setting a comparison parameter $D$, and comparing $D$ with the euclidean distance $d_i, i \in \mathbb{N}$ of two near and not directly closed (i.e.: a row or column apart), edge calculated pints $a_{\sigma1}$ and $a_{\sigma2}$. If  $D$ is large than $d_i$, directly connect points $a_{\sigma1}$ and $a_{\sigma2}$ as the new edge line segment. After gaining the edge of the object, the next step is removing the edges lines that are unable to form an effective closed loop, have a too small length, or have a chaotic, discrete distribution. The remaining edges lines are the satisfied ones. Finally, combining with the image area segmentation, removing the background more effective. 
	
\begin{figure}[t]
		\centering
		\includegraphics[width=0.5\columnwidth]{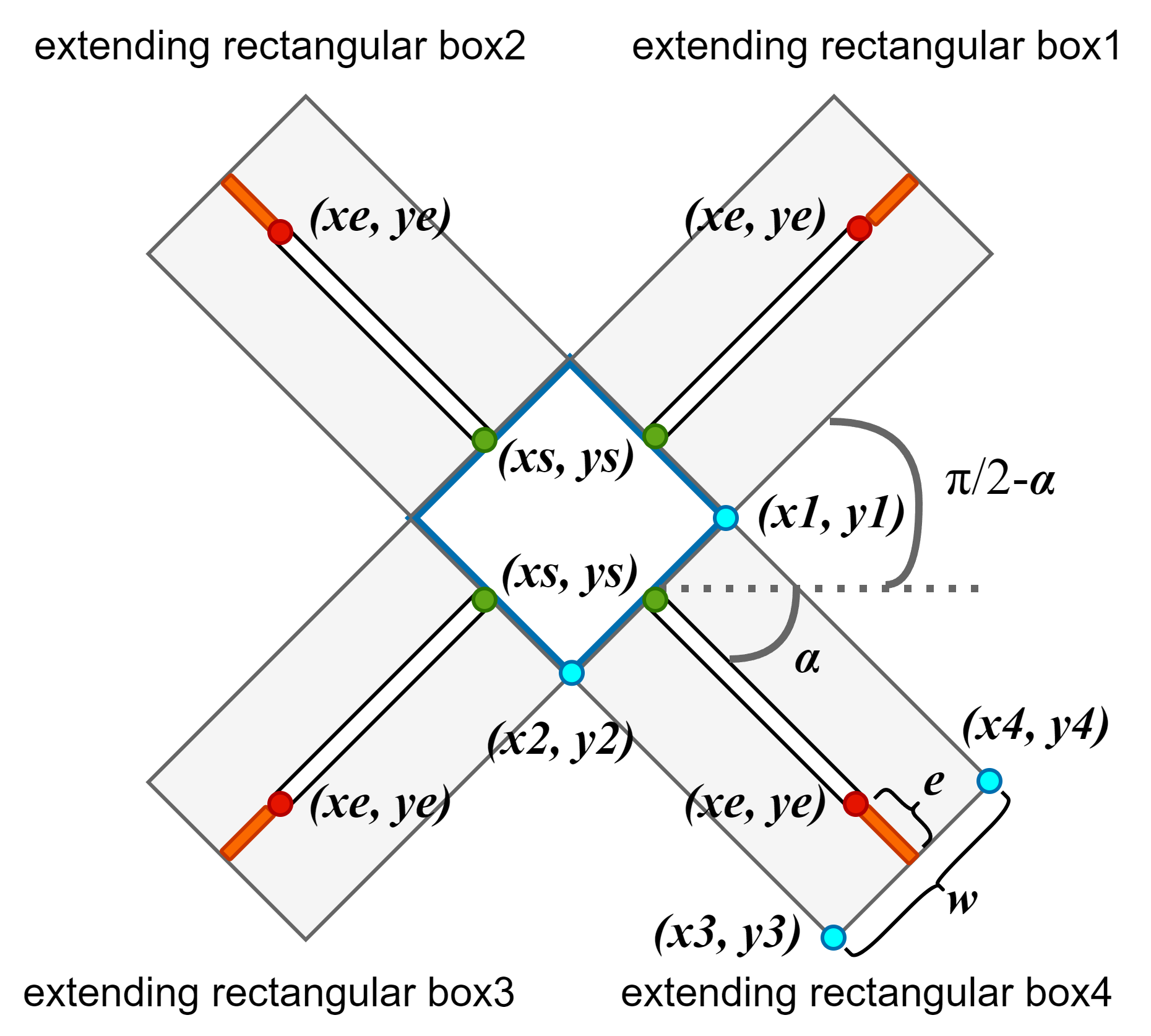}
		\caption{Removing the background with trunk area and extending rectangular box.}
		\label{fig6}
	\end{figure}

	The second method to improve the effect of removing the background of the input image we proposed is using the trunk domains to assistant removing background, based on the key points detection of human body. The implementation procedure can be described as Equation \ref{eq3}. As shown in Figure \ref{fig6}, the parameter $\alpha$ represents the anticlockwise radian value between the trunk line segment $(x_s, y_s) \to (x_e, y_e)$, which uses the point $(x_s, y_s)$ represents the start point and point $(x_e, y_e)$ represents the end point. The parameter $w$ represents the width of the extending rectangular box, the parameter $e$ represents the extended length following along the line direction of line segment $(x_s, y_s) \to (x_e, y_e)$ of the extending rectangular box. After gaining the coordinates of the extending rectangular box $((x_1, y_1), (x_2, y_2), (x_3, y_3), (x_4, y_4))$ for one bone segment $(x_s, y_s) \to (x_e, y_e)$, in the similar way, calculating the all gaining the coordinates for all of the bones of the basic armature in key-points. Note that the detected key-points of the object basic armature may have some excursions, e.g.: using the MMPOSE is hard to get the right key-points of the object head part for some model image. For this issue, a solution is define a enough large extending rectangular box with the neck points as the start point and a enough length for the rectangular box that is usually 3-6 times length of the distance from neck point to mouth. Meanwhile, the problem of the extending rectangular box is out of input original image area should be also considered.

	\begin{equation}
		\label{eq3}
		\left\{
		\begin{aligned}
			(x_i, y_i) = (x_s \pm \frac{w}{2cos(\frac{\pi}{2}-\alpha)},  y_s \pm \frac{w}{2sin(\frac{\pi}{2}-\alpha)}), \ i \in [1, 2]   \\
			(x_j, y_j) = ((x_e \pm \frac{e}{cos(\alpha)}) \pm \frac{w}{2cos(\frac{\pi}{2}-\alpha)},  (y_e \mp \frac{e}{cos(\alpha)}) \pm \frac{w}{2sin(\frac{\pi}{2}-\alpha)}), \ j \in [3, 4]   \\
			if (0 \le \alpha \le \pi) \to \alpha = -\alpha, else \to \alpha = \alpha, \alpha \in [0, 2\pi]
		\end{aligned}
		\right.
	\end{equation}

	The methods for removing the background mentioned above are independent and can be make used of by combing use. Method 1 is the basic, method 2 and method 3 can improve the performance effect of removing the background of the input image. Besides removing the background, in the pre-processing stage of OnTo3D, various of other methods can be used to improve the effect, e.g.: for the blurred image, using the deblurring methods with machine learning; for the image with a object that locates not in the center, capturing the valid semantic area in the middle of the image; for the image with light colors, improving the brightness of the color of the image, etc. After all of the pre-processing of the input image, the image can be used to generate a better 3D model compared without the pre-processing.

	\subsection{Armature}
	
	Generating the self-adaption armature is the one of the important parts of this version of OneTo3D. In this section, we will make the statement and analyzing of basic construction of the whole armature. Due to the basic processing object type is human-like, for explain the basic design idea of this armature more simply, we designed a special basic armature to control main parts of the human object body trunks. 
	
\begin{figure}[h]
		\centering
		\begin{minipage}{0.49\linewidth}
			\centering
			\includegraphics[width=0.9\linewidth]{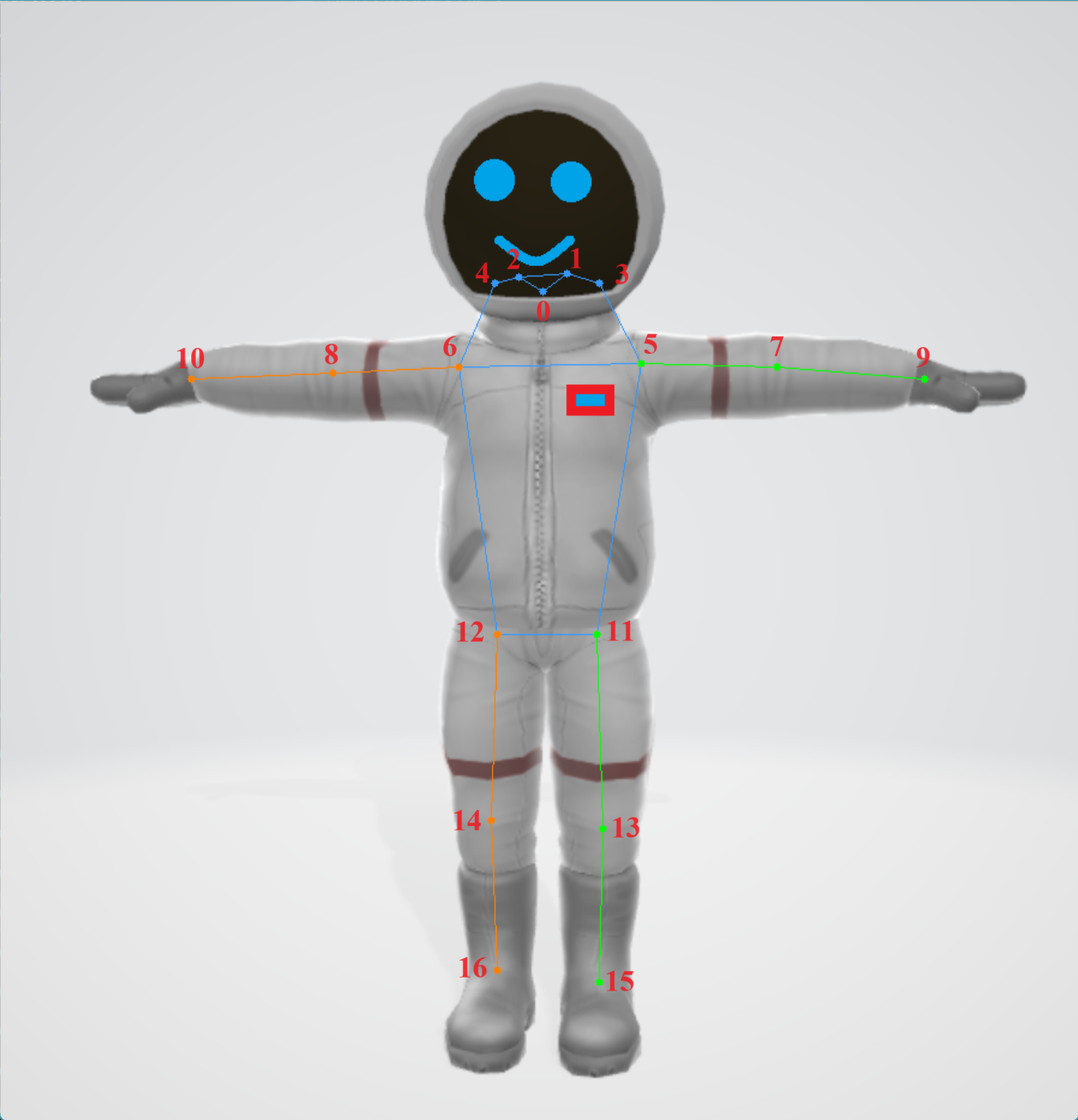}
			\centerline{(a) key points indexes of the armature.}
		\end{minipage}
		\begin{minipage}{0.49\linewidth}
			\centering
			\includegraphics[width=0.9\linewidth]{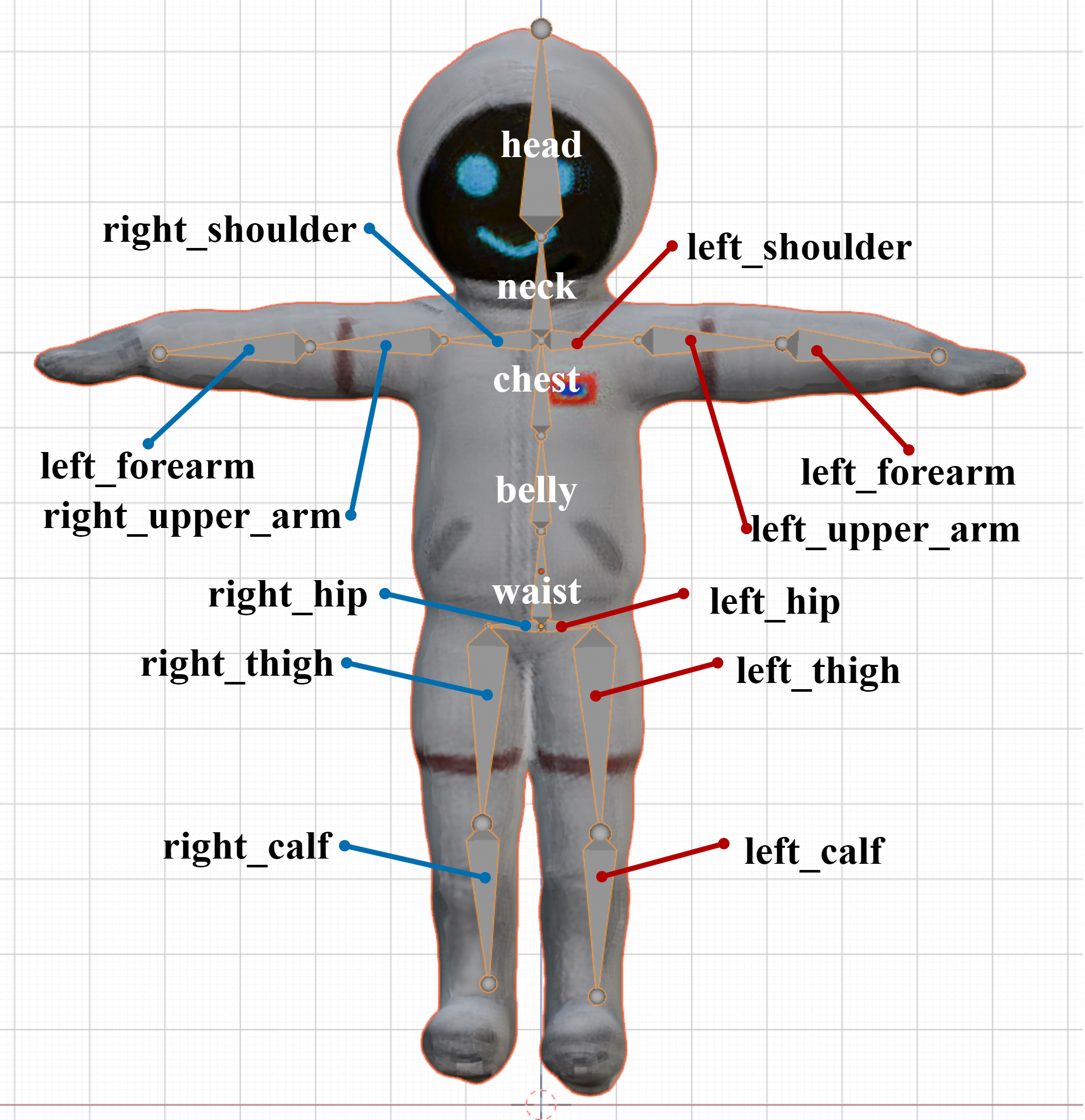}
			\centerline{(b) name IDs of the bones of the armature.}
		\end{minipage}
		\qquad
		\begin{minipage}{0.49\linewidth}
			\centering
			\includegraphics[width=0.9\linewidth]{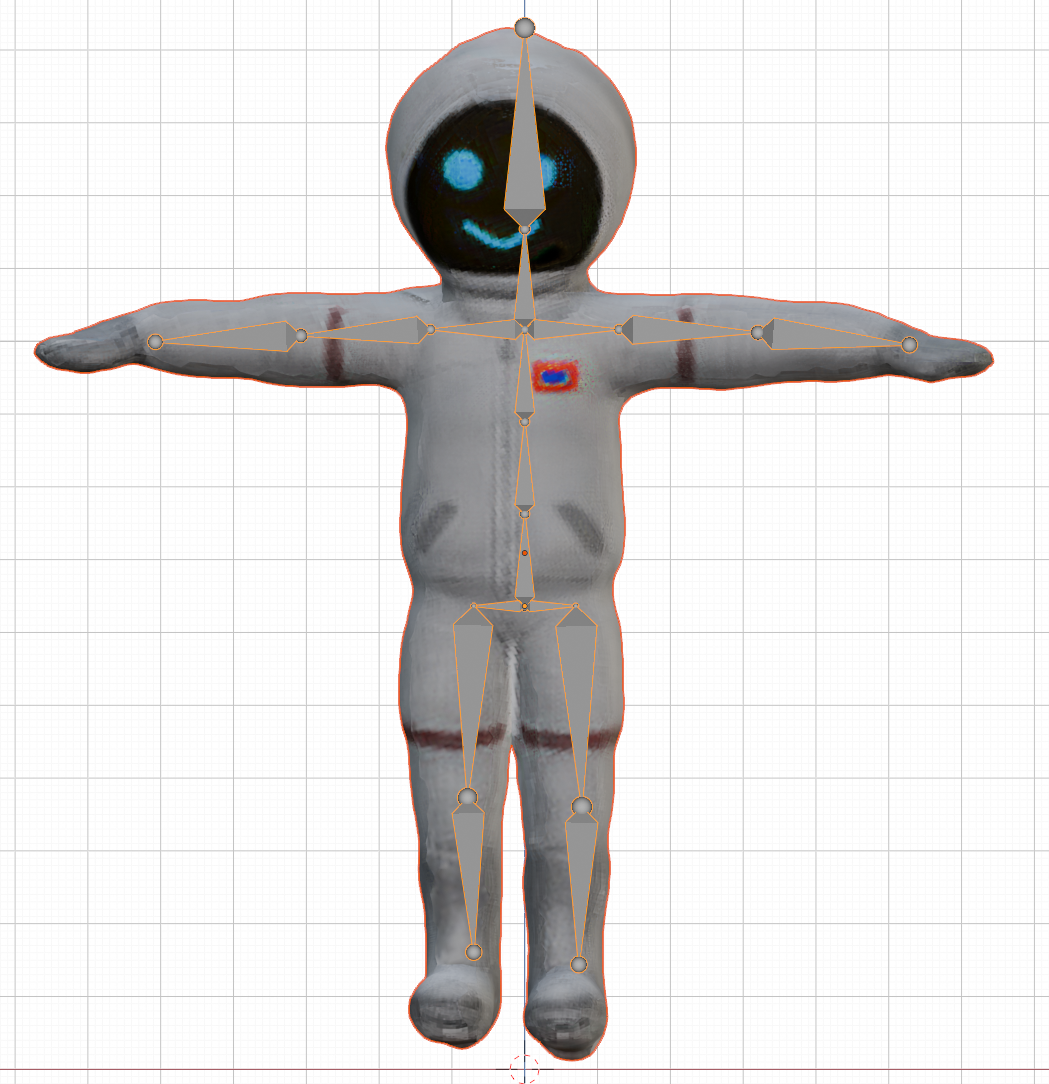}
			\centerline{(c) generated armature in 2D projection.}
		\end{minipage}
		\begin{minipage}{0.49\linewidth}
			\centering
			\includegraphics[width=0.9\linewidth]{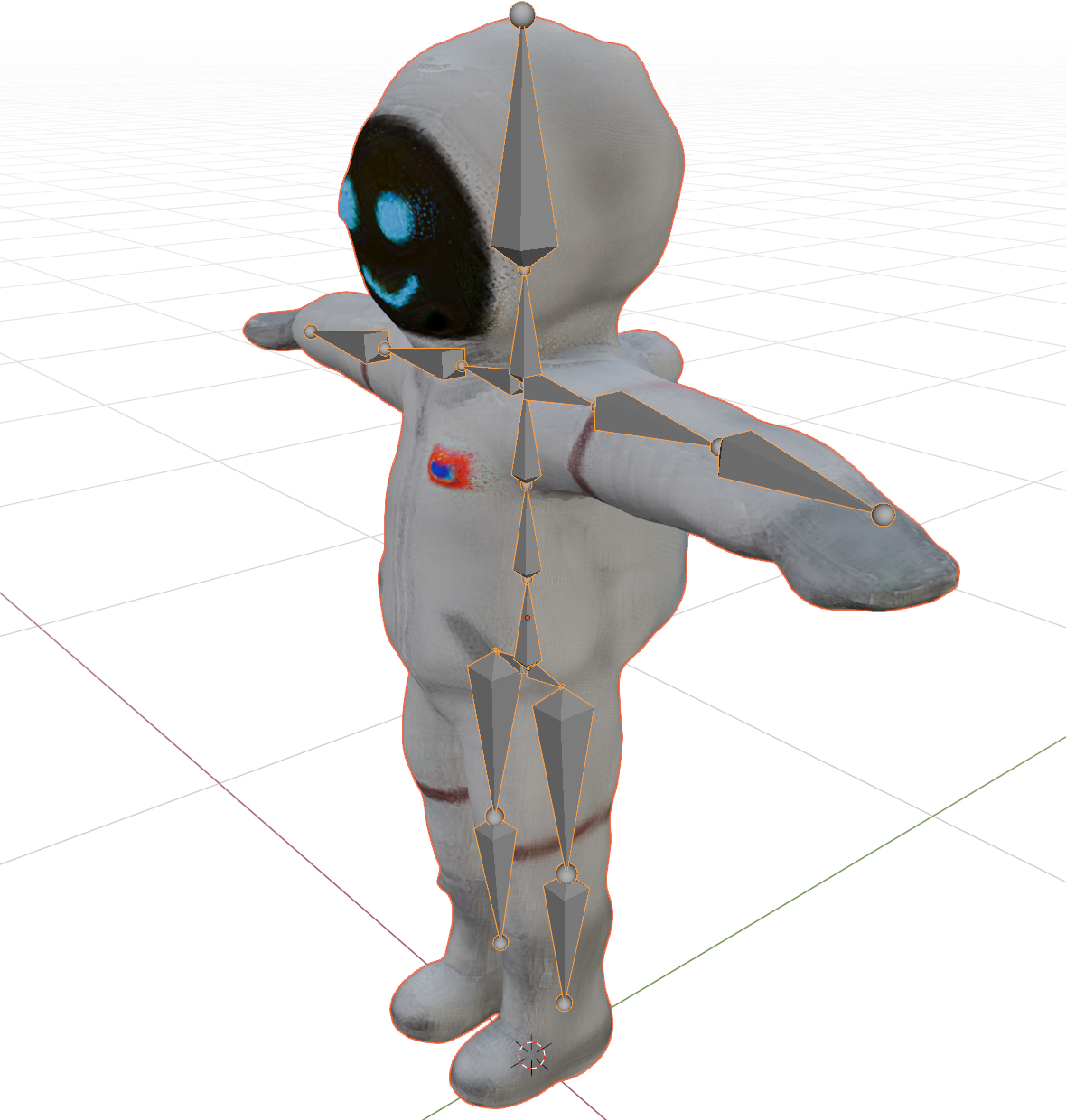}
			\centerline{(d) generated armature in 3D projection.}
		\end{minipage}
		
		\caption{Removing the background with edge detection. }
		\label{fig7}
	\end{figure}

	The precision of the generating the basic armature is positively related to the precision of the location of key-points. As shown in Figure\ref{fig7}-(a), the key-points of the object are gained by MMPOSE. After enough experiments, we found that using the MMPOSE is hard to detect the key-points of the parts on face, which may because of the complexity and diversity of the abstract human object face. Therefore, the key-points of the object gained from OneTo3D can not be used in the generation of basic 3D armature to generate the head or neck bone, or the further self-adaption adjustment. However for the generation of other bones, the key-points is useful. 
	
	For controlling the bones of the basic armature more easily, each bone of the armature is named with a special name identity(ID). As shown in Figure\ref{fig7}-(b), the names are: head, neck, chest, belly, waist, left\_shoulder, left\_forearm, left\_upper\_arm, left\_hip, left\_thigh, left\_calf, right\_shoulder, right\_forearm, right\_upper\_arm, right\_hip, right\_thigh, right\_calf. The generating order is from waist to head, from components in left part to right part, from father node bones to children node bones. The implementation of bones creation is make use of the $bpy.ops.armature.extrude\_move$ function of blender to extrude the children node bone from the last father level node bone. There are five father-to-children segmentation in the generated armature: from waist to head, from left\_shoulder to left\_forearm, from right\_shoulder to right\_forearm, from left\_hip to left\_calf, from right\_hip to right\_calf. The order is same as the generating order. After the self-adaption adjustment, the generated armature will fit the volume of the object, as shown in Figure\ref{fig7}-(c) and Figure\ref{fig7}-(d). More details about the self-adaption armature will be presented in following section.

	\subsection{Self-adaption}
	
	Self-adaption adjustment is the core idea of generating a suitable armature for the targeted object, which is one of the most important part of this version of OneTo3D. Directly using the key-points from MMPOSE to generate the armature will usually causes the mismatch between the 3D object model and 3D armature, which mainly be expressed in the location and size scaling of the armature.  
	
\begin{figure}[t]
		\centering
		\includegraphics[width=0.99\columnwidth]{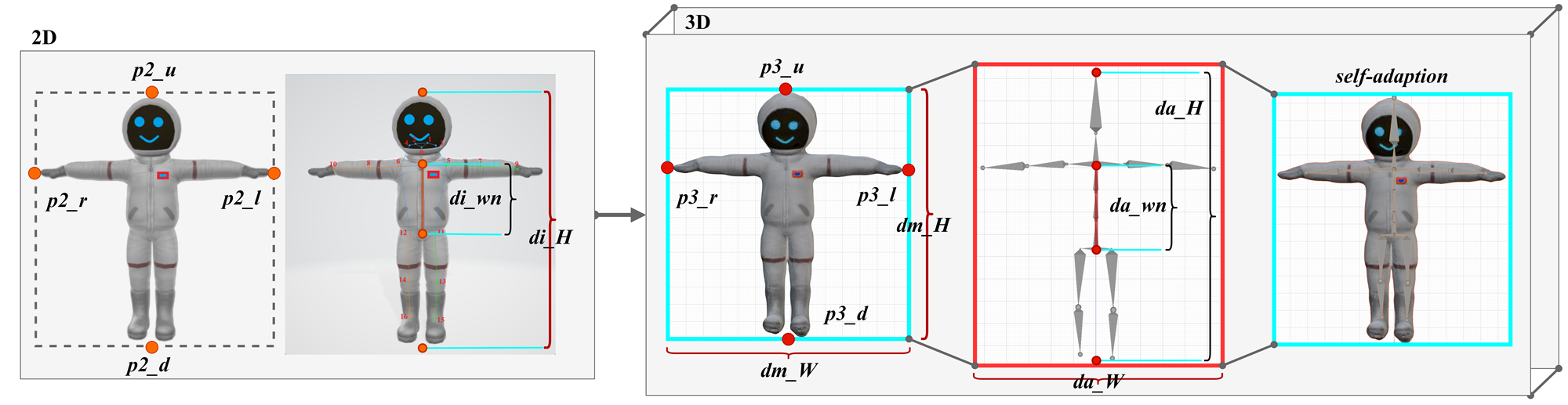}
		\caption{Self-adaption mechanism of generating the armature based on ratio scaling and parameter adjustment.}
		\label{fig8}
	\end{figure}

	Self-adaption adjustment is the core idea of generating a suitable armature for the targeted object, which is one of the most important part of this version of OneTo3D. Directly using the key-points from MMPOSE to generate the armature will usually causes the mismatch between the 3D object model and 3D armature, which mainly be expressed in the location and size scaling of the armature. Due to there are some relative ratio relationship between the key-points detected and the generate armature, it is able to design the self-adjustment of the generated armature to make the armature fit the generated 3D model. As shown in Figure\ref{fig8}, after the structure analyzing, we selected two segment distances for the ratio scaling calculation, i.e.: the segment between the points $waist$ and $neck$ in the 2D image, which is represented as $di_{wn}$; and the segment distance between the bones start points $waist$ and $neck$, which is represented as $da_{wn}$. Using the underscores to indicate subscripts. 
	
\begin{figure}[t]
		\centering
		\includegraphics[width=0.98\columnwidth]{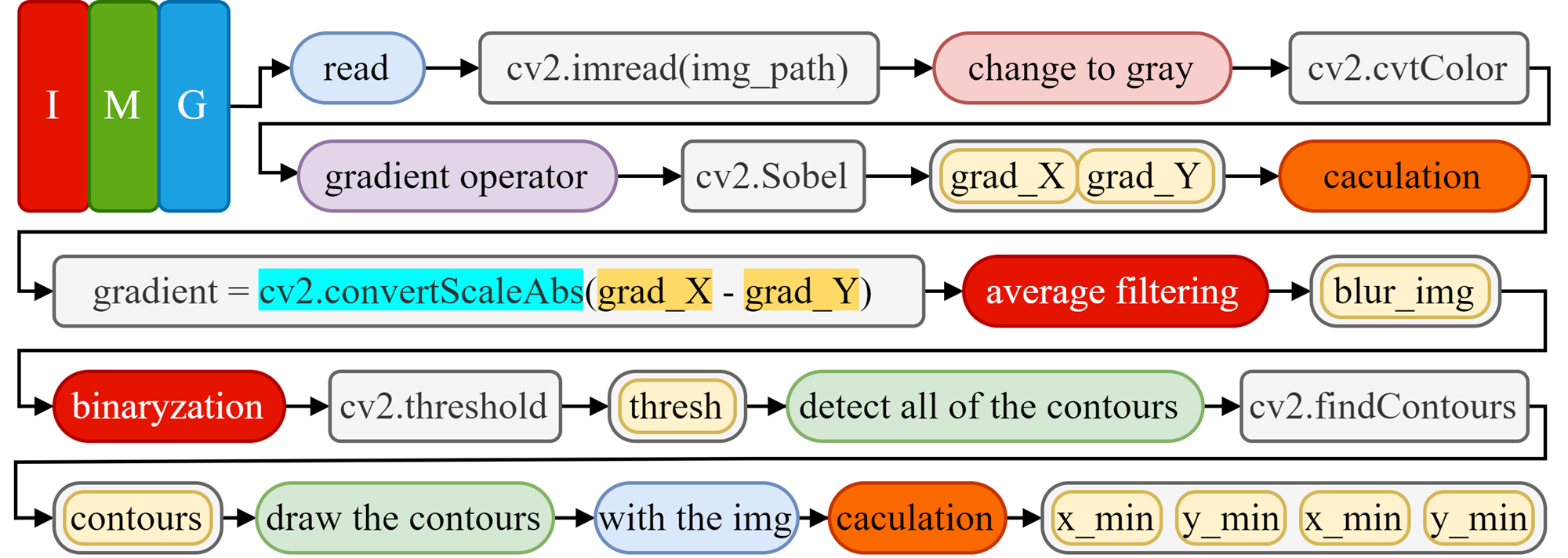}
		\caption{Gaining the boundary points based on function $cv2.findContours$ and optimal value judgment.}
		\label{fig9}
	\end{figure}

	We proposed two methods to address this issue. As shown in Equation\ref{eq4}, the finial goal is to gain the high of the armature being generated, which is represented as $da_H$. As shown in Figure \ref{fig8}, using the points $p2_u, p2_d, p2_l, p2_r$ to represent the representative boundary points in the directions up, down, left and right respectively. Meanwhile, using the points $p3_u, p3_d, p3_l, p3_r$ to represent the representative boundary points of 3D model projection on the front view in the directions up, down, left and right respectively. Due the all components of the 3D model is generated in equal proportions with the 2D input image as reference, it only need to take one component geometrical distance in 2D as reference, with the equal proportions, to calculate all of the component geometrical distances of the 3D model, as well as the component geometrical distances of the model armature. Using $di_H$ to present the high of the object in 2D input image. Variable $dm_H$ represents the high of the object in 3D model. Using $di_H$ to calculate the $dm_H$ that is closed to $da_H$. As shown in Equation\ref{eq4}, to gain the $di_H$ and $dm_H$, the first step is to gain the coordinates of $p2_u, p2_d, p2_l, p2_r$ which are represented with $(x_{2u}, y_{2u}), (x_{2p}, y_{2p}, (x_{2l}, y_{2u})$ respectively, and the coordinates of $p2_u, p2_d, p2_l, p2_r$ which are represented with $(x_{2u}, y_{2u}), (x_{2p}, y_{2p}, (x_{2l}, y_{2u})$ respectively. To gain the coordinates of $p2_u, p2_d, p2_l, p2_r$, we designed our detection algorithm based on the $cv2.drawContours$ of OpenCV-Python. The main idea is acquiring all of the possible contours of the image with function $cv2.findContours$, as shown in Figure \ref{fig9}, the first step is reading and changing the RGB input image to gray image. Subsequently, calculating the gradient changing following the directions $X$ and $Y$ respectively, which are represented by $grad\_X$ and $grad\_Y$ respectively. Following calculating the gradient value $gradient$, using the function $cv2.blur$ to make the average filtering operation and get result image $blur\_img$. Then making the binaryzation for the image $blur\_img$. Implementing the threshold processing with specific threshold values and get result $thresh$. Eventually, using the function $cv2.findContours$ to find all of the possible contours $contours$. The following operation of drawing the contours with the input image is optional. Subsequently, calculating the contours boundary points coordinates as four points $(x_{min}, y_{min}), (x_{min}, y_{max}), (x_{max}, y_{min}), (x_{max}, y_{max})$, which can be used to calculate the $P_{im}$ and $dm_H$ as shown in Equation \ref{eq4}.
	
	\begin{equation}
		\label{eq4}
		\left\{
		\begin{aligned}
			[p2_u, p2_d, p2_l, p2_r] = [(x_{2u}, y_{2u}), (x_{2d}, y_{2d}), (x_{2l}, y_{2l}), (x_{2r}, y_{2r})]   \\
			[p3_u, p3_d, p3_l, p3_r] = [(x_{3u}, y_{3u}), (x_{3d}, y_{3d}), (x_{3l}, y_{3l}), (x_{3r}, y_{3r})]   \\
			P_{im} = \frac{di_H}{dm_H} =  \frac{di_H}{da_H} = \frac{|(y_{2u}-y_{2d})|}{|(y_{3u}-y_{3d})|} = \frac{|(y_{2l}-y_{2r})|}{|(y_{3l}-y_{3r})|}   \\
			\frac{|(y_{2u}-y_{2d})|}{|(y_{3u}-y_{3d})|} = \frac{|(y_{max}-y_{min})|}{|(y_{3u}-y_{3d})|};\ \frac{|(y_{2l}-y_{2r})|}{|(y_{3l}-y_{3r})|} = \frac{|(x_{max}-x_{min})|}{|(y_{3l}-y_{3r})|}   \\
		\end{aligned}
		\right.
	\end{equation}
	
	\begin{equation}
		\label{eq5}
		\left\{
		\begin{aligned}
			P_{im} = \frac{da_{wn}}{di_{wn}}\\
			da_{s_b \to e_b} = dm_{s_b \to e_b} = P_{im} * di_{s_b \to e_b},\ b \in Bones   \\
			Ra_{s_b \to e_b} = Rm_{s_b \to e_b} = Ri_{s_b \to e_b},\ b \in Rotations   \\
			da_H = dm_H = |(y_{3u}-y_{3d})|;\ da_W = dm_W = |(y_{3l}-y_{3r})|   \\
			di = di_{s_b \to e_b},\ b \in Bones \cap b \notin [neck, head]   \\
			di_{neck} = p_{neck} \cdot dori_{(neck \to mouth)};\ di_{head} = p_{head} \cdot dori_{(neck \to mouth)}   \\
		\end{aligned}   
		\right.
	\end{equation}
	
	The second method to make the self-adaption generation of the armature is designing the bones lengths of armature with the information of key-points detection. As shown in Figure\ref{fig8}, due the distance between the start pints of components $waist$ and $neck$ is usually changeless, we selected this bones segment distance as the basic reference of the scaling calculation. The detailed main calculation is shown in Equation \ref{eq5}. Using the $P_{im}$ to represent the scaling ratio of from the being generated armature to the input image, which used the distance between the start pints of components $waist$ and $neck$ as basic reference. Set $Bones$ as the bone names or indexes set. $da_{s_b \to e_b}$ represents the bone $b$ distance ratio (i.e.: bone length ratio compared with bone segment $waist \to neck$) of armature, correspondingly, $dm_{s_b \to e_b}$ represents the model space distance ratio of bone $b$ inside the model; $di_{s_b \to e_b}$ represents the 2D projection distance ratio of the bone $b$. Set $Rotations$ as the bone rotation degrees set. $Ra_{s_b \to e_b}$ represents the bone $b$ rotation, correspondingly, $Rm_{s_b \to e_b}$ represents the model space rotation degree of bone $b$ inside the model; $di_{s_b \to e_b}$ represents the 2D projection rotation degree of the bone $b$. Due the equal scaling between the generated armature and input image key-points, these rotation degrees is equal. Parameter $da_H$ represents the high of the generated armature and $da_H$ represents the high of the 3D model, which are equal. Parameter $da_W$ represents the high of the generated armature and $da_W$ represents the width of the 3D model, which are equal. Parameter $d_i$ represents the distance ratio of bone $i$ in $Bones$ but not includes the bones $neck$ and $head$, because of the key-points of the bone nodes grained from MMPOSE has unstable precision in the detection of head and face components. For the bone length detection of bones $neck$ and $head$, we defined two scaling parameters $p_{neck}$ and $p_{head}$ to scale the distance ratio from key point $neck$ to key point $mouth$ that is named as $dori_{(neck \to mouth)}$ to define the distance ratio of bone $neck$ and $head$. With the value of $di$, the value of $Pim$ can be calculated. Further, using $da_{s_b \to e_b} = P_{im} * di_{s_b \to e_b}$ to gain each length of bone segment $da_{s_b \to e_b}$.
	
	To guarantee the bones generation of the generated 3D model armature appropriately, based on the parameter $P_{im}$ gained from Equations\ref{eq4} and \ref{eq5}, we designed an algorithm for calculating the next point coordinate of the parameter $value$ of function $bpy.ops.armature.extrude\_move$ in the definition for next bone extruding.

	\begin{equation}
		\label{eq6}
		\left\{
		\begin{aligned}
			X = (endP[0] - startP[0]) \cdot P_{im}   \\
			Y = (endP[1] - startP[1]) \cdot P_{im}   \\
			if: direction = left \to value = (abs(X), 0, Y)   \\
			elif: direction = right \to value = (-abs(X), 0, Y)   \\
		\end{aligned}   
		\right.
	\end{equation}

	When implementing the self-adaption between the armature and the 3D model, gaining the bound box data (i.e.: the size data of the 3D model volume) and the space location data are significant. As shown in Equation\ref{eq7}, using the $bpy.context.object$ to locate and represent the generated 3D model and is abbreviated as $obj$. Using $obj.bound\_box$ to gain the bounding box data from Blender. There are 8 bound points in $bbox$, from where we selected two points $bbox[0]$ and $bbox[6]$ to implement the calculation. Parameters $X$, $Y$, $Z$ represent the dimension lengths of the generated 3D model in Blender editing space from the directions $z, y, z$ respectively. It is also available to gain the dimension lengths by using the $bpy.context.object.dimensions[d], d \in [0 \to x, 1 \to y, 2 \to z]$. After gaining the data of the $bbox$, it is feasible to calculate the center dimension lengths from the directions $z, y, z$ respectively, which are represented by $X_{center}, Y_{center}, Z_{center}$ respectively. Re-locating the location of the generated 3D model will be significant for the self-adaption armature binding quality. Therefore, we use the $obj.location$ for fine tuning the re-localization of the 3D model. The fine tuning parameters $\tau_{x}, \tau_{y}, \tau_{z}, \in \mathbb{Q}$ represent the distance fine tuning in positive or negative direction.

	\begin{equation}
		\label{eq7}
		\left\{
		\begin{aligned}
			obj = bpy.context.object  \to bbox = obj.bound\_box   \\
			X = (bbox[6][0] - bbox[0][0])  \cdot  obj.scale.x   \\
			Z = (bbox[6][1] - bbox[0][1])  \cdot  obj.scale.z   \\
			Y = (bbox[6][2] - bbox[0][2])  \cdot  obj.scale.y   \\
			X_{center} = \frac{(bbox[6][0] + bbox[0][0])}{2}  \cdot  obj.scale.x   \\
			Y_{center} = \frac{(bbox[6][2] + bbox[0][2])}{2}  \cdot  obj.scale.y   \\
			Z_{center} = \frac{(bbox[6][1] + bbox[0][1])}{2}  \cdot  obj.scale.z   \\
			obj.location = (X_{center} + \tau_{x}, Y_{center} + \tau_{y}, \frac{Z}{2} + \tau_{z})
		\end{aligned}   
		\right.
	\end{equation}
	
	With our enough experiments, we found that the basic 3D models generated from Zero123 model are not always locate in one same position, which means the generation self-adaption armature should consider about the fine tuning of the miss-distance. From most of the 3D models that have geometric center point in middle of the model image projection, the self-adaption armature will gain a good binding with the 3D model. Otherwise, fine tuning the value of $\tau_{x}$, $\tau_{y}$ or $\tau_{z}$ from $0.01 \to 0.05$ to gain a good binding in the directions of $z, y, z$ respectively. Using the $\frac{Z}{2}$ can make the positive displacement of 3D model along the Z axis, which will make the view angle of final generation of the re-editable 3D model or 3D video better. Parameters $obj.scale.x$, $obj.scale.y$ and $obj.scale.z$ represent the scaling factors of the 3D model in the directions of $z, y, z$ respectively. More detailed implementations are presented in project code.

	\subsection{Interpreter}
	
\begin{figure}[t]
		\centering
		\includegraphics[width=0.8\columnwidth]{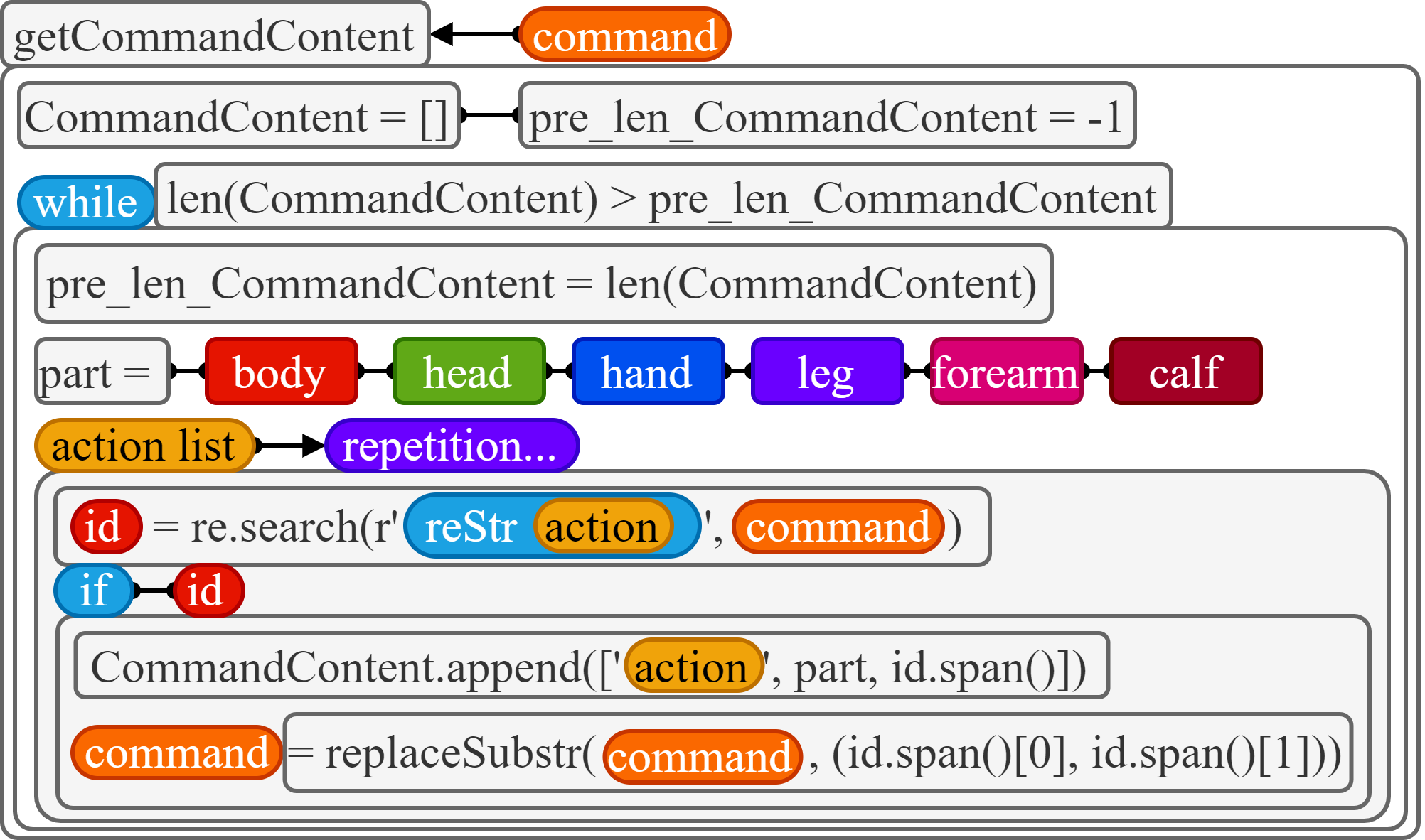}
		\caption{Interpreter mechanism for input command to gain command control detail.}
		\label{fig10}
	\end{figure}

	After the achievement of the generation of self-adaption armature, the next step is interpreting the user input command and gain the command control detail, which is called the interpreter of OneTo3D. As shown in Figure\ref{fig10}, the main idea of interpreting the command detail is implemented in function $getCommandContent$, with the user input command $command$ as input variable. Using the $pre\_len\_CommandContent$ as a judgment benchmark for the main while loop. The user input command text can be a long text that contain multiple action control detail information. Each action control detail will be split and understanding before being saved in list $CommandContent$. We designed a while loop with the comparison of length of list $CommandContent$ and the number of read command items as loop judgment condition. The function of the while loop is to read all of the sub-command details that area contained in the user input command text $command$, which is also the core of this interpreter.   
	
\begin{figure}[t]
		\centering
		\includegraphics[width=1\columnwidth]{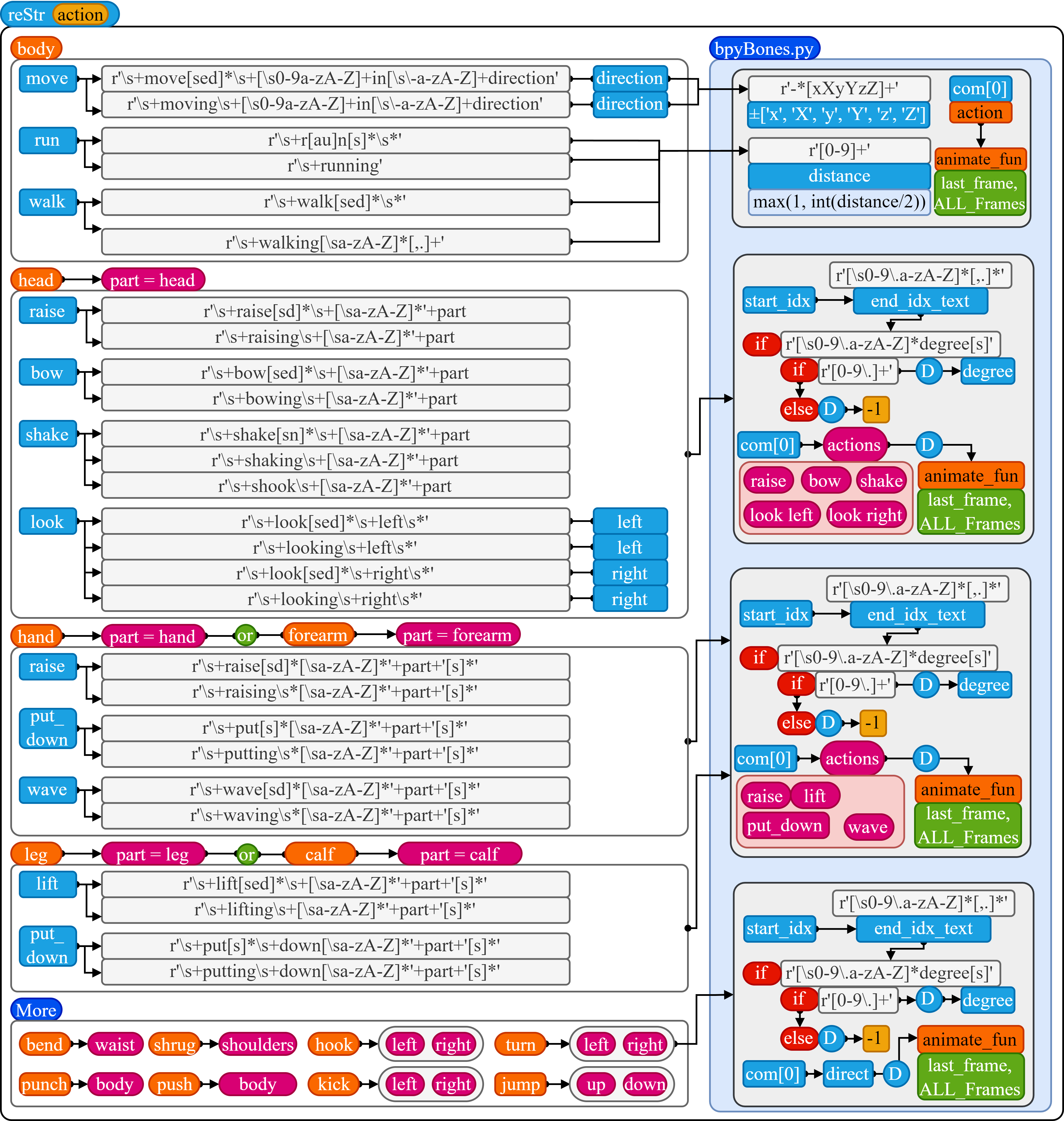}
		\caption{Interpreter with the mechanism of special actions judgment regular expressions structures.}
		\label{fig11}
	\end{figure}

	In the while loop, the first processing is updating the iteration adding value of $pre\_len\_CommandContent$. Subsequently, defining the variable part to represent one body component of the generated 3D model. For easily controlling the 3D model to make the simple actions, we defined five main body parts to be defined, i.e.: body, head, hand, leg, forearm and calf. Following, enter the processing of action detail interpretation of body part command. Various of action detail extraction and judgment will be implemented based on regular expressions. As shown in Figure\ref{fig10}, one user input $command$ text may contain various of sub-command actions, which are considered as an action list. Applying coding a similar repetition that contains multiple $if$ actions judgment regular expressions structures, to make the action extraction and detail judge for all of the supported actions. The input command text that contains various of action details and regular expressions structures, is represented as $reStr$. As shown in Figure\ref{fig11}, we designed a special interpreter mechanism for the interpretation of $reStr$. There are six main action parts in this version of OneTo3D, i.e.: $body$, $head$, $hand$, $forearm$, $leg$, $calf$. The action interpretation for part $hand$ is similar to part $forearm$, while the action interpretation for part $leg$ is similar to part $calf$. We also provided more possible supported action in the code, e.g.: the actions of turning left and right. The while loop ensures that all of the action data can be read. The session of $reStr$ interpreting will be implemented in code file $animation.py$, the whole result will be save in a text file $command.txt$, and called to read data by the code file $bpyBones.py$ for further processing, which can be shown in Figure\ref{fig11}. 
	
	Return to Figure\ref{fig10}, for each possible supported action issues, we design a special regular expressions structures to extract the action detail information. And the action detail searching and extracting return result will be defined as $id$. If the return result is not null, adding the new action extracted data of action, action implementation part, regular expression matching information into $CommandContent$ list. After recording the current sub-text action data, it is needed to update the current $command$ variable with the default method that replacing the analyzed action sub-text with multiple special characters which will not confuse the rest of action information (e.g.: dash $-$). When recording sub-actions data, $id.span()[0]$ record the action implementing part, $id.span()[1]$ record the start index and end index of the corresponding sub-command text in the original command. The corresponding sub-command text of the current sub-action may contain the quantization words (e.g.: directions or number).  
	
	With enough experiments, we found that it is infeasible to run the Blender to implement the armature and actions rendering with Python function parameters directly. Therefore, after gaining the $CommandContent$ list, the interpreted actions data will save into a text file $command.txt$ writing line by line. Subsequently, we designed the further interpreter mechanism, which is implemented in code file $bpyBones.py$, as shown in Figure\ref{fig11}, for the command list to extract the details of the a series of actions in user command. We used a for loop to make the sub-action judgment for each possible condition. For each sub-action item $com$ in command list (i.e.: $CommandContent$ list read from $command.txt$), judging the action type that is included in $com[0]$, or judging the action part that is included in $com[1]$. For actions:$move$, $run$ and $walk$, judging the action type with $com[0]$, which is implemented in action part $body$. For the motion (e.g.: move) with direction (e.g.:$\pm x$, $\pm y$, $\pm z$), using a regular expressions $r'-*[xXyYzZ]+'$ to search and extract the detailed data of direction. If it is needed to add detailed distances, times or steps data in motion commands and interpret them, using a regular expressions $r'[0-9]+'$ to search and extract the detailed motion data (e.g.: for motions: $walk$ and $run$).
	
	For actions $turn\ left$ and $turn\ right$, judging the action type with $com[0]$, which will analyze and interpret the action degree. For the interpretation of the action degree in this mechanism, first defining the start index of the current being searched action sub-string and named as $start\_idx=com[2][0]$, then using a regular expressions $r'[\backslash s0-9\backslash.a-zA-Z]*[,.]*'$ to search and extract the main data containing sub-string further from the processed filtered original input command text $command\_words[start\_idx:]$ ($command\_words$ represents the original command text string). Subsequently, defining a default degree that is presented as $D$, with a default value 360°(covers more perspectives than 0°) that will not impact the final design and rendering result. Following, using a regular expressions $r'[\backslash sa-zA-Z]*[0-9\backslash.]+[\backslash sa-zA-Z]*degree[s]'$ to extract the sub-string that contains the degree quantization information. If the searched sub-string includes degree quantization information, then using the regular expressions $r'[0-9\backslash.]+'$ to extract the detailed number of the degree that supports floating-point format. If there is not degree quantization information in the sub-string, set $D$ as -1, which means the degree will use default value. Following animate function (i.e.: $animate\_turn$) that is represented as $animate\_fun$, will control the corresponding bones of the generated armature to make commanded action or motion, including adding the key-frame in Blender for 3D video generation. 
	
	For the action part $head$, judging $com[1]$, similarly, first judging and gaining the action degree with the method mentioned above and gain the action degree $D$. Subsequently, recognizing the action contain in $com[0]$. For action part $head$, the supported actions include: raise, bow, shake, look left, look right. The corresponding animation functions $animate\_fun$ are: $animate\_raise\_head$, $animate\_bow\_head$, $animate\_shake\_head$, $animate\_look\_left$ and $animate\_look\_right$. The action animation function $animate\_fun$ will return two parameters: $last\_frame$ that represents the last processed frame index, $ALL_Frames$ that represents the count of processed frame, which are able to be sent to next action animation function $animate\_fun$ for the continuous action and motion design and generation. For action parts hand, forearm, leg and calf, they used the similar processing algorithm design because of the similar kinematics structure. For action parts $hand$ and $forearm$, we designed the actions: $raise$, $put\_down$ and $wave$. For action parts $leg$ and $calf$, we designed the actions: $lift$ and $put\_down$. All of these four action parts support the customized side configuration $left$ or $right$ and the action rotation degree $D$. When there is not explicit content about the action side in the user input command text of one action that should be defined with action side, OneTo3D will select one action side from left or right randomly. The interpreter of this version of OneTo3D supports multiple-actions combined command.

	\subsection{3D Video}
	
	\begin{figure}[t]
		\centering
		\includegraphics[width=1\columnwidth]{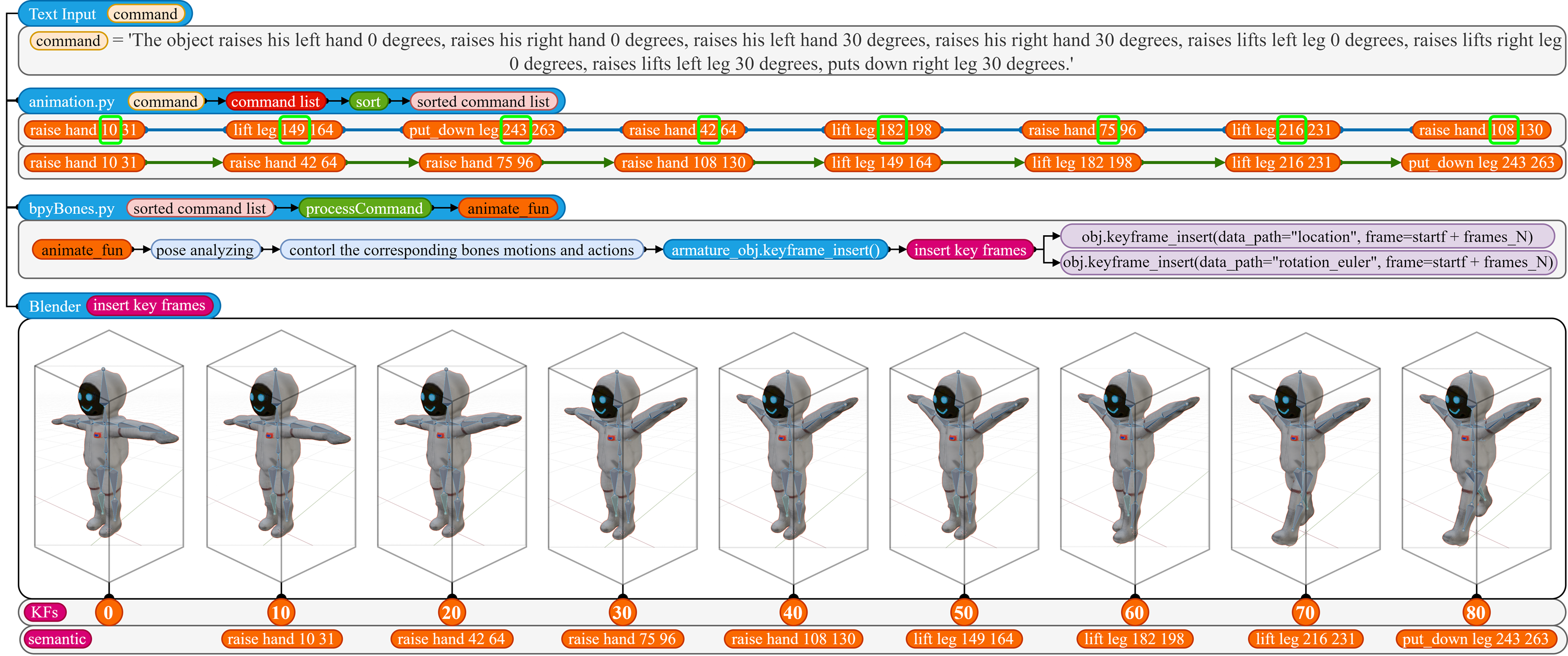}
		\caption{Generating the 3D video with armature controlling of generated 3D model and Blender.}
		\label{fig12}
	\end{figure}
	
	In the design of this version of OneTo3D, we used the Blender as the basic assisted 3D video development and generation toolkit. As shown in Figure\ref{fig12}, the user input command text is shown as follow: 
	
	\begin{verbatim}
		The object raises his left hand 0 degrees, raises his right hand 0 degrees, 
		raises his left hand 30 degrees, raises his right hand 30 degrees, 
		raises lifts left leg 0 degrees, raises lifts right leg 0 degrees, 
		raises lifts left leg 30 degrees, puts down right leg 30 degrees.
	\end{verbatim}
	
	When generating the 3D video, the first step is using the interpreter mechanism to analyze the input command text and gain the corresponding command list. For the interpreter, the action verbs in command are tense-independent. As shown in Figure\ref{fig12}, with the design statements mentioned before, the interpreter mechanism is designed and divided into two main parts: being implemented in code files $animation.py$ and $bpyBones.py$ respectively. In the processing part in $animation.py$, the interpreter will search and gain the sub-commands and sort them in a command list. The sub-commands in the command list follow one same data structure: $[action\_type, action\_part, sub\_command\_start\_index, \\ sub\_command\_end\_index]$. For example, for the unsorted first sub-command $[raise\ hand\ 10\ 31]$, in which the parameter $raise$ represents the $action\_type$, $hand$ means $action\_part$, $10$ is the $sub\_command\_start\_index$ (i.e.: the start index of sub-command sub-string in original command string) and $31$ expresses the $sub\_command\_end\_index$ (i.e.: the end index of sub-command sub-string in original command string), which means the corresponding sub-command is:$'\ raises\ his\ left\ hand'$. Sorting the sub-commands in the command list uses the position with the index value of 2 as the sort reference basic position, that means sorting the $sub\_command\_start\_index$ to sort sub-commands.      
	
	Subsequently, the second part of the interpreter designed in code file $bpyBones.py$ will read and analyzed the sorted $command\ list$ with the implementation of $processCommand$ functions. Following, each action sub-command will be interpreted to detailed action data in the corresponding action analyzing function $animate\_fun$, in where the pose detail contained in the action data will be extracted and analyzed. Subsequently, the 3D action controlling mechanism will control the corresponding bones to implement the targeted motions and actions. After the relative bones of armature have implemented the targeted motions and actions, using the function $armature\_obj.keyframe\_insert()$ to insert the current pose of the 3D model as a key-frame. For the actions of coordinates motion, defining the $data\_path$ as $location$, while for the actions of rotation actions, defining the $data\_path$ as $rotation\_euler$. Finally, as shown in Figure\ref{fig12}, calling the Blender as an generation assistant, completing the inserting of all of the action key-frames $KFs$, and generating a whole 3D video (the 3D Blender re-editable file can be generated meanwhile). Parameter $semantic$ represents the corresponding relationship between the key-frames and the sub-commands in the $command\ list$. The generation function will not remember or generate the action of key-frame 0 of the 3D model, therefore an available method to record the initial pose of some relative bones, when designing 3D video generation with coding of Blender, is making a sub-command with no motion or action change(e.g.:with 0 degree rotation).

	\section{Evaluation}
	
	In this section, we will analyze the real rendering and generation abilities of this version of OneTo3D. In most of the comparison items and evaluation criterion, OneTo3D can gain an eminent and splendid performance.
	
	\subsection{Self-adaption and Generation}
	
\begin{figure}[h]
		\centering
		
		\begin{minipage}{0.09\linewidth}
			\centering
			\includegraphics[width=0.9\linewidth]{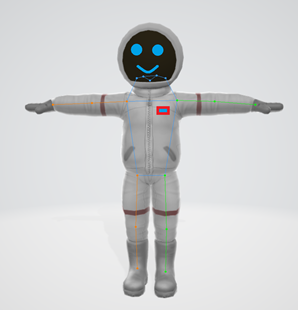}
		\end{minipage}
		\begin{minipage}{0.09\linewidth}
			\centering
			\includegraphics[width=0.9\linewidth]{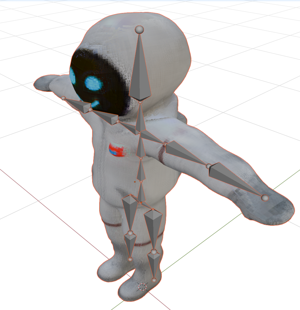}
		\end{minipage}
		\begin{minipage}{0.09\linewidth}
			\centering
			\includegraphics[width=0.9\linewidth]{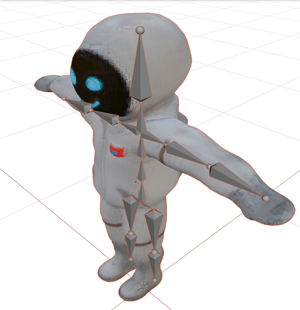}
		\end{minipage}
		\begin{minipage}{0.09\linewidth}
			\centering
			\includegraphics[width=0.9\linewidth]{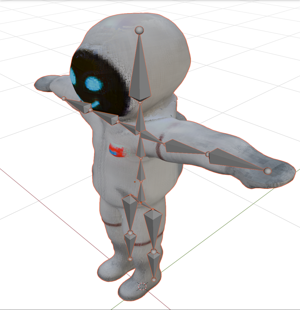}
		\end{minipage}
		\begin{minipage}{0.09\linewidth}
			\centering
			\includegraphics[width=0.9\linewidth]{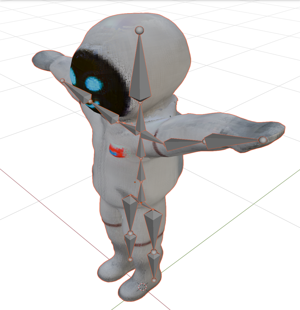}
		\end{minipage}
		\begin{minipage}{0.09\linewidth}
			\centering
			\includegraphics[width=0.9\linewidth]{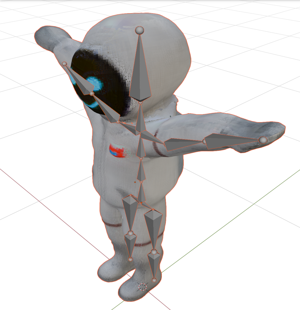}
		\end{minipage}
		\begin{minipage}{0.09\linewidth}
			\centering
			\includegraphics[width=0.9\linewidth]{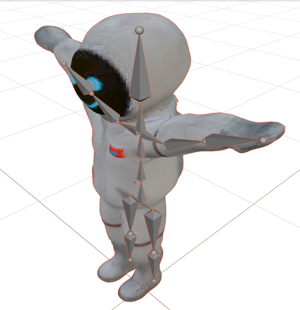}
		\end{minipage}
		\begin{minipage}{0.09\linewidth}
			\centering
			\includegraphics[width=0.9\linewidth]{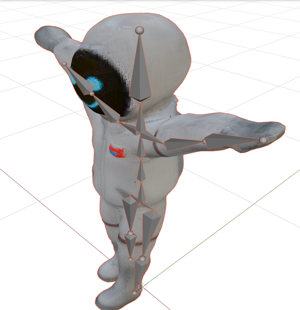}
		\end{minipage}
		\begin{minipage}{0.09\linewidth}
			\centering
			\includegraphics[width=0.9\linewidth]{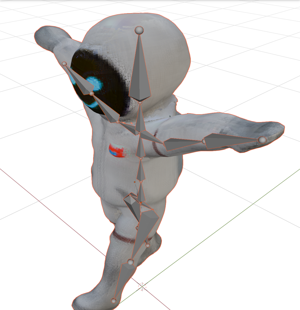}
		\end{minipage}
		\begin{minipage}{0.09\linewidth}
			\centering
			\includegraphics[width=0.9\linewidth]{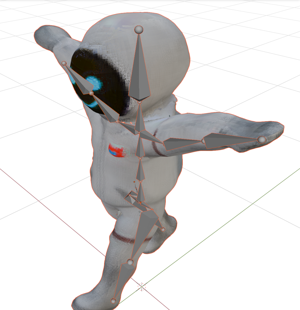}
		\end{minipage}
		\qquad

		\begin{minipage}{0.09\linewidth}
			\centering
			\includegraphics[width=0.4\linewidth]{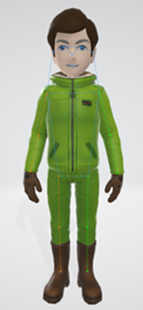}
		\end{minipage}
		\begin{minipage}{0.09\linewidth}
			\centering
			\includegraphics[width=0.9\linewidth]{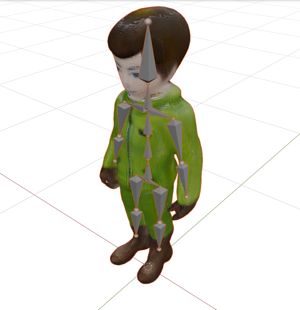}
		\end{minipage}
		\begin{minipage}{0.09\linewidth}
			\centering
			\includegraphics[width=0.9\linewidth]{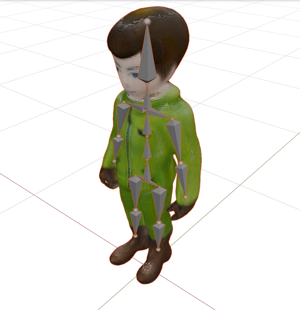}
		\end{minipage}
		\begin{minipage}{0.09\linewidth}
			\centering
			\includegraphics[width=0.9\linewidth]{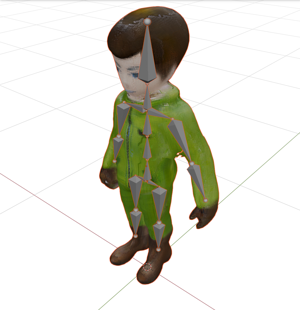}
		\end{minipage}
		\begin{minipage}{0.09\linewidth}
			\centering
			\includegraphics[width=0.9\linewidth]{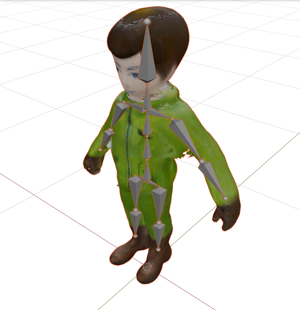}
		\end{minipage}
		\begin{minipage}{0.09\linewidth}
			\centering
			\includegraphics[width=0.9\linewidth]{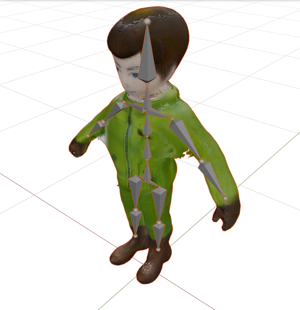}
		\end{minipage}
		\begin{minipage}{0.09\linewidth}
			\centering
			\includegraphics[width=0.9\linewidth]{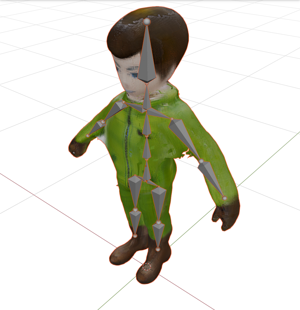}
		\end{minipage}
		\begin{minipage}{0.09\linewidth}
			\centering
			\includegraphics[width=0.9\linewidth]{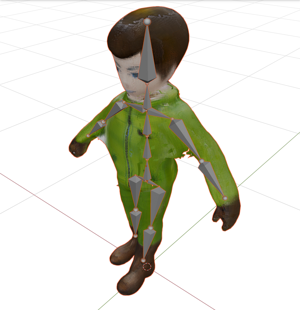}
		\end{minipage}
		\begin{minipage}{0.09\linewidth}
			\centering
			\includegraphics[width=0.9\linewidth]{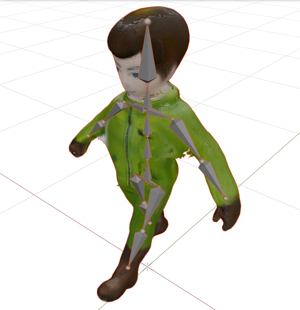}
		\end{minipage}
		\begin{minipage}{0.09\linewidth}
			\centering
			\includegraphics[width=0.9\linewidth]{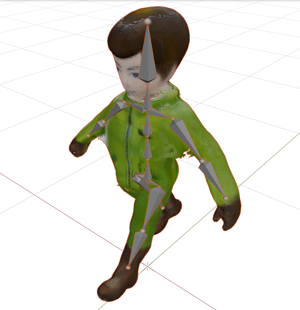}
		\end{minipage}
		\qquad

		\begin{minipage}{0.09\linewidth}
			\centering
			\includegraphics[width=0.9\linewidth]{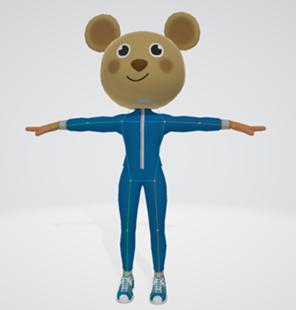}
		\end{minipage}
		\begin{minipage}{0.09\linewidth}
			\centering
			\includegraphics[width=0.9\linewidth]{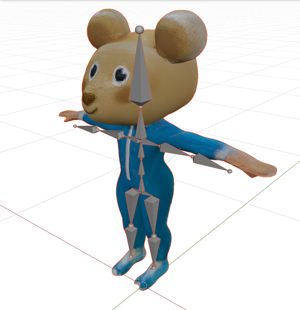}
		\end{minipage}
		\begin{minipage}{0.09\linewidth}
			\centering
			\includegraphics[width=0.9\linewidth]{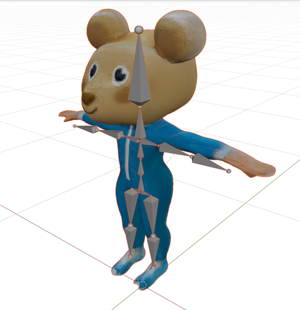}
		\end{minipage}
		\begin{minipage}{0.09\linewidth}
			\centering
			\includegraphics[width=0.9\linewidth]{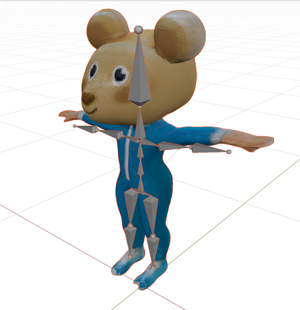}
		\end{minipage}
		\begin{minipage}{0.09\linewidth}
			\centering
			\includegraphics[width=0.9\linewidth]{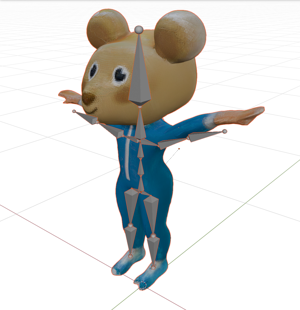}
		\end{minipage}
		\begin{minipage}{0.09\linewidth}
			\centering
			\includegraphics[width=0.9\linewidth]{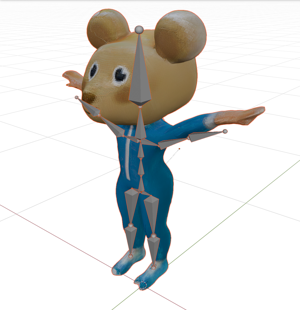}
		\end{minipage}
		\begin{minipage}{0.09\linewidth}
			\centering
			\includegraphics[width=0.9\linewidth]{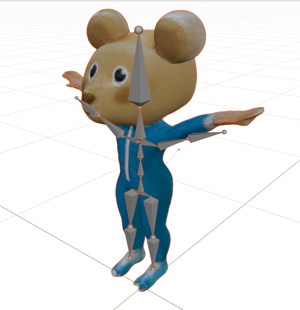}
		\end{minipage}
		\begin{minipage}{0.09\linewidth}
			\centering
			\includegraphics[width=0.9\linewidth]{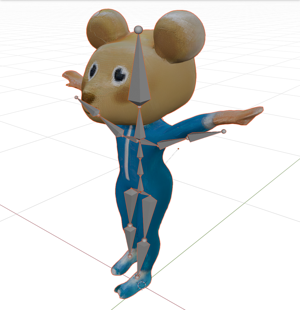}
		\end{minipage}
		\begin{minipage}{0.09\linewidth}
			\centering
			\includegraphics[width=0.9\linewidth]{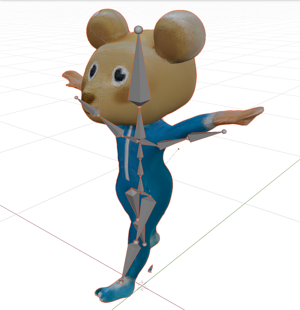}
		\end{minipage}
		\begin{minipage}{0.09\linewidth}
			\centering
			\includegraphics[width=0.9\linewidth]{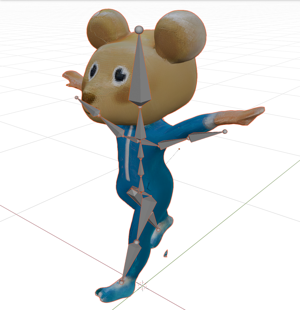}
		\end{minipage}
		\qquad

		\begin{minipage}{0.09\linewidth}
			\centering
			\includegraphics[width=0.56\linewidth]{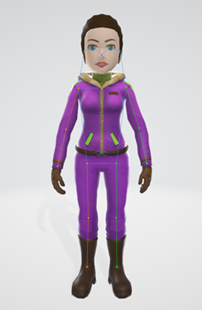}
		\end{minipage}
		\begin{minipage}{0.09\linewidth}
			\centering
			\includegraphics[width=0.9\linewidth]{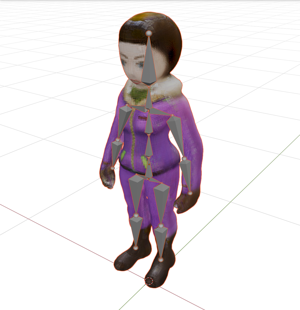}
		\end{minipage}
		\begin{minipage}{0.09\linewidth}
			\centering
			\includegraphics[width=0.9\linewidth]{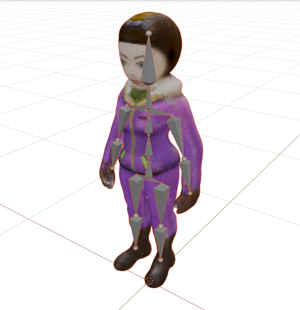}
		\end{minipage}
		\begin{minipage}{0.09\linewidth}
			\centering
			\includegraphics[width=0.9\linewidth]{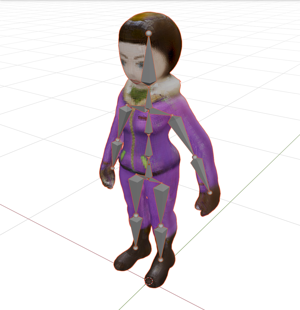}
		\end{minipage}
		\begin{minipage}{0.09\linewidth}
			\centering
			\includegraphics[width=0.9\linewidth]{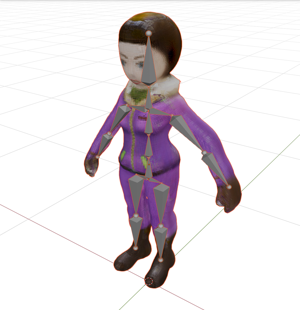}
		\end{minipage}
		\begin{minipage}{0.09\linewidth}
			\centering
			\includegraphics[width=0.9\linewidth]{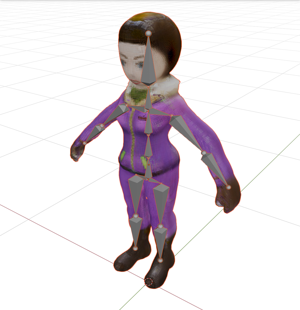}
		\end{minipage}
		\begin{minipage}{0.09\linewidth}
			\centering
			\includegraphics[width=0.9\linewidth]{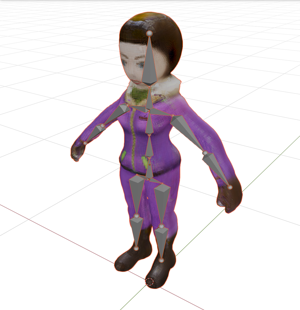}
		\end{minipage}
		\begin{minipage}{0.09\linewidth}
			\centering
			\includegraphics[width=0.9\linewidth]{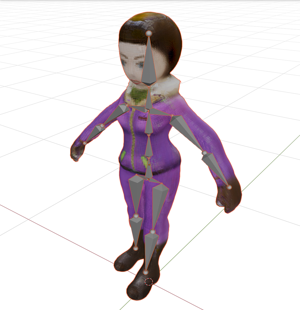}
		\end{minipage}
		\begin{minipage}{0.09\linewidth}
			\centering
			\includegraphics[width=0.9\linewidth]{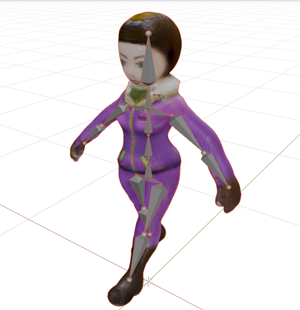}
		\end{minipage}
		\begin{minipage}{0.09\linewidth}
			\centering
			\includegraphics[width=0.9\linewidth]{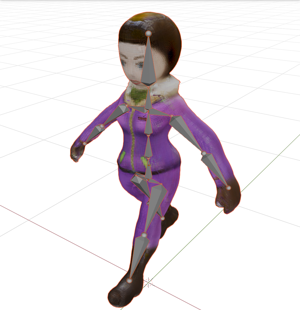}
		\end{minipage}
		\qquad

		\begin{minipage}{0.09\linewidth}
			\centering
			\includegraphics[width=0.5\linewidth]{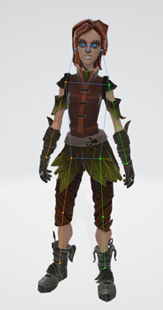}
		\end{minipage}
		\begin{minipage}{0.09\linewidth}
			\centering
			\includegraphics[width=0.9\linewidth]{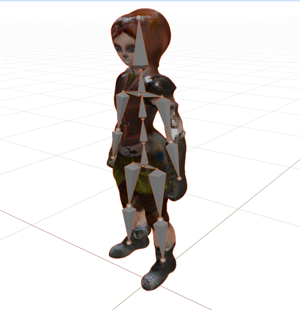}
		\end{minipage}
		\begin{minipage}{0.09\linewidth}
			\centering
			\includegraphics[width=0.9\linewidth]{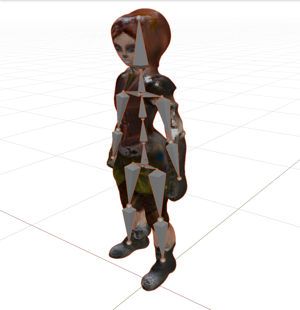}
		\end{minipage}
		\begin{minipage}{0.09\linewidth}
			\centering
			\includegraphics[width=0.9\linewidth]{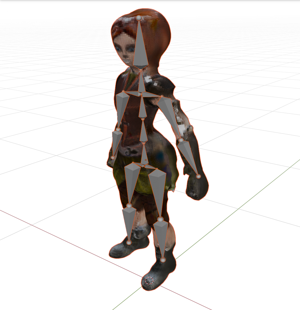}
		\end{minipage}
		\begin{minipage}{0.09\linewidth}
			\centering
			\includegraphics[width=0.9\linewidth]{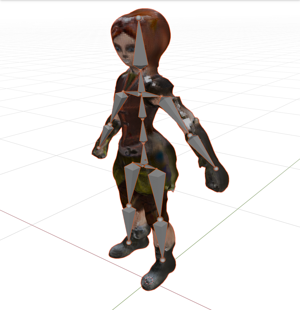}
		\end{minipage}
		\begin{minipage}{0.09\linewidth}
			\centering
			\includegraphics[width=0.9\linewidth]{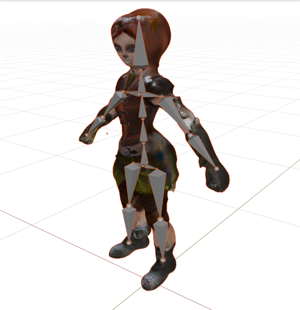}
		\end{minipage}
		\begin{minipage}{0.09\linewidth}
			\centering
			\includegraphics[width=0.9\linewidth]{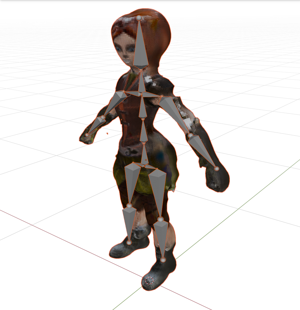}
		\end{minipage}
		\begin{minipage}{0.09\linewidth}
			\centering
			\includegraphics[width=0.9\linewidth]{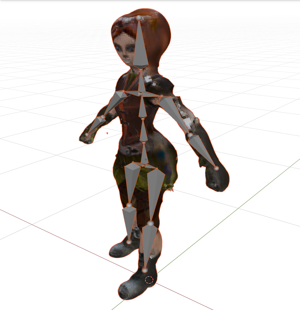}
		\end{minipage}
		\begin{minipage}{0.09\linewidth}
			\centering
			\includegraphics[width=0.9\linewidth]{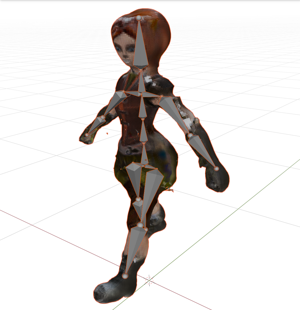}
		\end{minipage}
		\begin{minipage}{0.09\linewidth}
			\centering
			\includegraphics[width=0.9\linewidth]{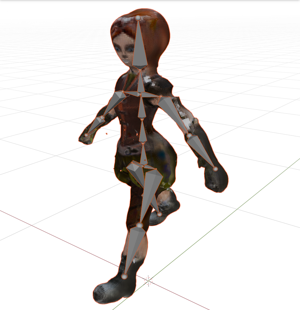}
		\end{minipage}
		\qquad

		\begin{minipage}{0.09\linewidth}
			\centering
			\includegraphics[width=0.8\linewidth]{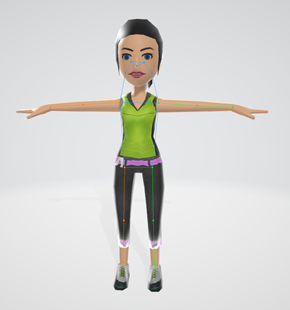}
		\end{minipage}
		\begin{minipage}{0.09\linewidth}
			\centering
			\includegraphics[width=0.9\linewidth]{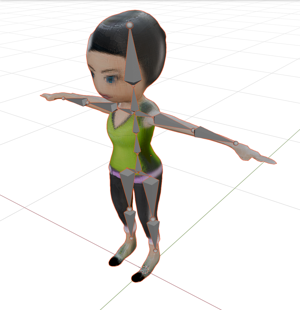}
		\end{minipage}
		\begin{minipage}{0.09\linewidth}
			\centering
			\includegraphics[width=0.9\linewidth]{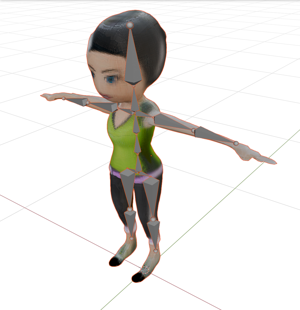}
		\end{minipage}
		\begin{minipage}{0.09\linewidth}
			\centering
			\includegraphics[width=0.9\linewidth]{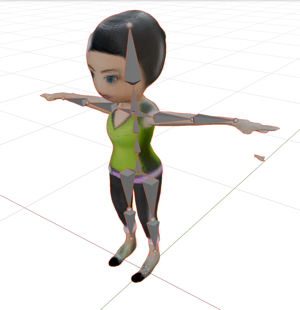}
		\end{minipage}
		\begin{minipage}{0.09\linewidth}
			\centering
			\includegraphics[width=0.9\linewidth]{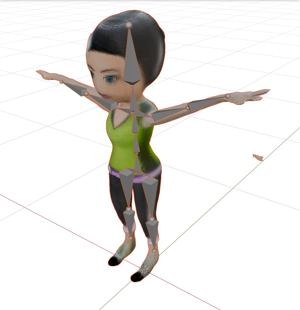}
		\end{minipage}
		\begin{minipage}{0.09\linewidth}
			\centering
			\includegraphics[width=0.9\linewidth]{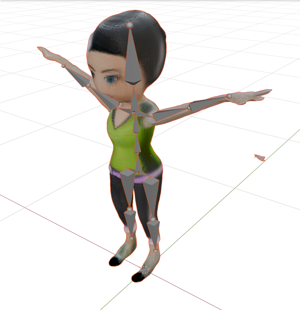}
		\end{minipage}
		\begin{minipage}{0.09\linewidth}
			\centering
			\includegraphics[width=0.9\linewidth]{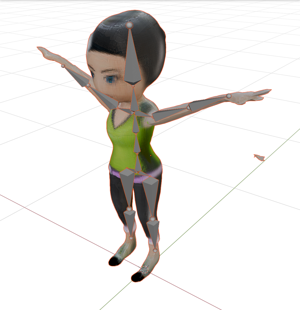}
		\end{minipage}
		\begin{minipage}{0.09\linewidth}
			\centering
			\includegraphics[width=0.9\linewidth]{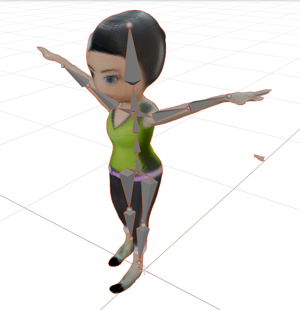}
		\end{minipage}
		\begin{minipage}{0.09\linewidth}
			\centering
			\includegraphics[width=0.9\linewidth]{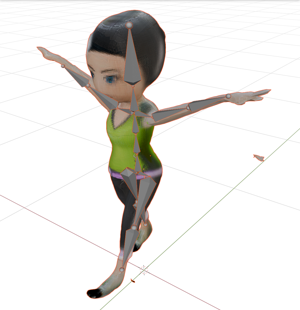}
		\end{minipage}
		\begin{minipage}{0.09\linewidth}
			\centering
			\includegraphics[width=0.9\linewidth]{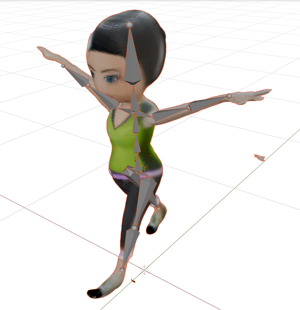}
		\end{minipage}
		\qquad

		\begin{minipage}{0.09\linewidth}
			\centering
			\includegraphics[width=0.45\linewidth]{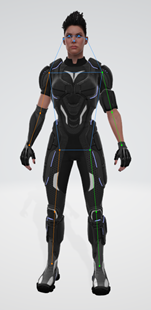}
		\end{minipage}
		\begin{minipage}{0.09\linewidth}
			\centering
			\includegraphics[width=0.9\linewidth]{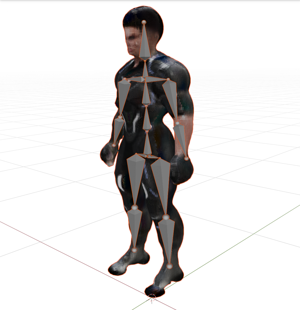}
		\end{minipage}
		\begin{minipage}{0.09\linewidth}
			\centering
			\includegraphics[width=0.9\linewidth]{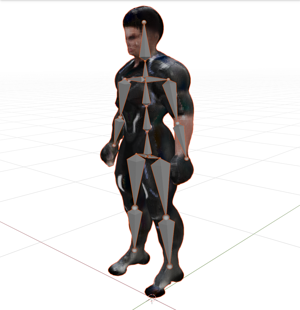}
		\end{minipage}
		\begin{minipage}{0.09\linewidth}
			\centering
			\includegraphics[width=0.9\linewidth]{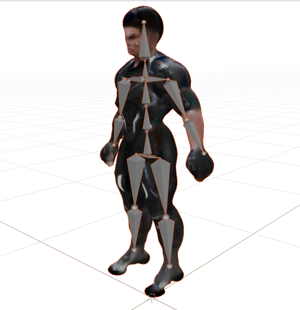}
		\end{minipage}
		\begin{minipage}{0.09\linewidth}
			\centering
			\includegraphics[width=0.9\linewidth]{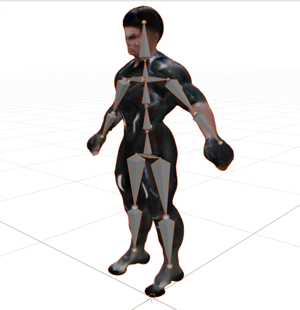}
		\end{minipage}
		\begin{minipage}{0.09\linewidth}
			\centering
			\includegraphics[width=0.9\linewidth]{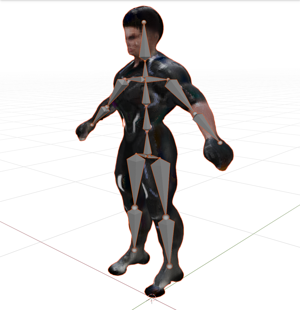}
		\end{minipage}
		\begin{minipage}{0.09\linewidth}
			\centering
			\includegraphics[width=0.9\linewidth]{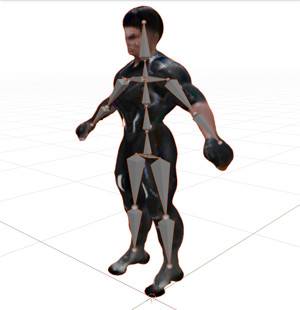}
		\end{minipage}
		\begin{minipage}{0.09\linewidth}
			\centering
			\includegraphics[width=0.9\linewidth]{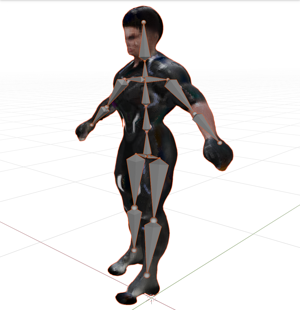}
		\end{minipage}
		\begin{minipage}{0.09\linewidth}
			\centering
			\includegraphics[width=0.9\linewidth]{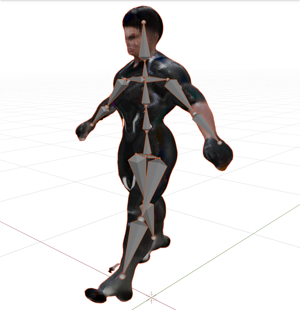}
		\end{minipage}
		\begin{minipage}{0.09\linewidth}
			\centering
			\includegraphics[width=0.9\linewidth]{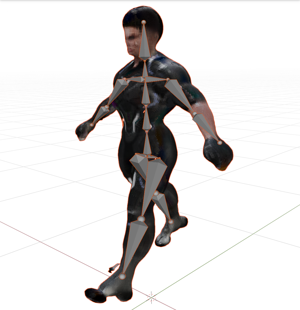}
		\end{minipage}
		\qquad

		\begin{minipage}{0.09\linewidth}
			\centering
			\includegraphics[width=0.45\linewidth]{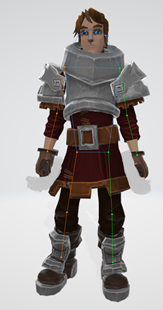}
		\end{minipage}
		\begin{minipage}{0.09\linewidth}
			\centering
			\includegraphics[width=0.9\linewidth]{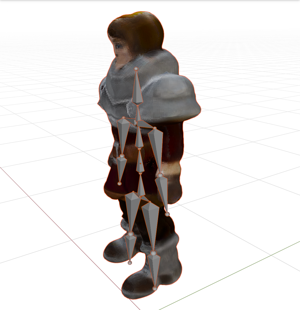}
		\end{minipage}
		\begin{minipage}{0.09\linewidth}
			\centering
			\includegraphics[width=0.9\linewidth]{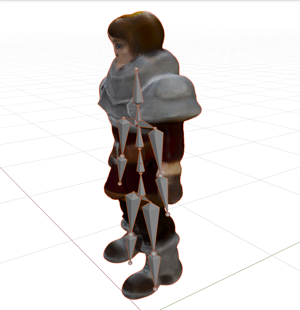}
		\end{minipage}
		\begin{minipage}{0.09\linewidth}
			\centering
			\includegraphics[width=0.9\linewidth]{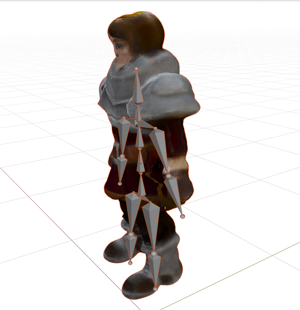}
		\end{minipage}
		\begin{minipage}{0.09\linewidth}
			\centering
			\includegraphics[width=0.9\linewidth]{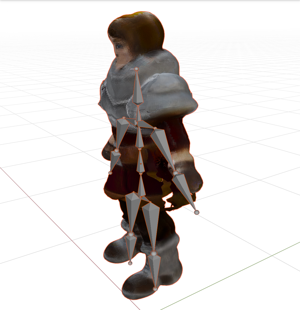}
		\end{minipage}
		\begin{minipage}{0.09\linewidth}
			\centering
			\includegraphics[width=0.9\linewidth]{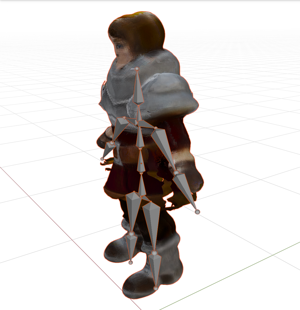}
		\end{minipage}
		\begin{minipage}{0.09\linewidth}
			\centering
			\includegraphics[width=0.9\linewidth]{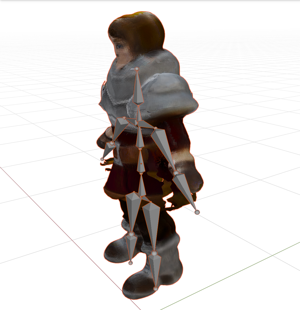}
		\end{minipage}
		\begin{minipage}{0.09\linewidth}
			\centering
			\includegraphics[width=0.9\linewidth]{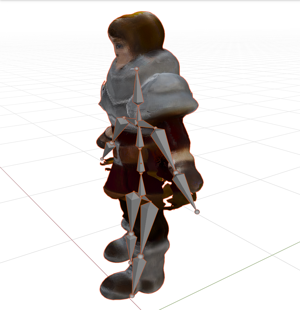}
		\end{minipage}
		\begin{minipage}{0.09\linewidth}
			\centering
			\includegraphics[width=0.9\linewidth]{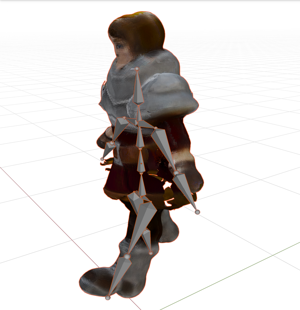}
		\end{minipage}
		\begin{minipage}{0.09\linewidth}
			\centering
			\includegraphics[width=0.9\linewidth]{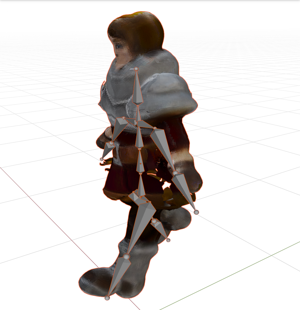}
		\end{minipage}
		\qquad

		\begin{minipage}{0.09\linewidth}
			\centering
			\includegraphics[width=0.9\linewidth]{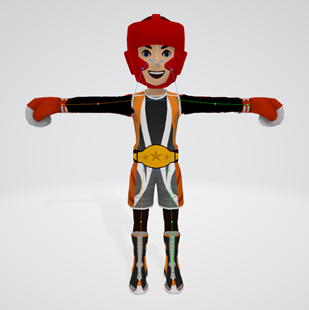}
			\centerline{image}
		\end{minipage}
		\begin{minipage}{0.09\linewidth}
			\centering
			\includegraphics[width=0.9\linewidth]{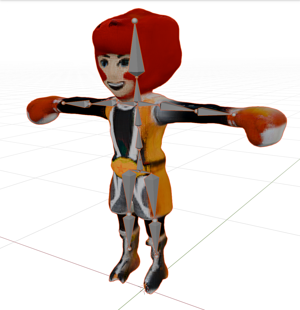}
			\centerline{0}
		\end{minipage}
		\begin{minipage}{0.09\linewidth}
			\centering
			\includegraphics[width=0.9\linewidth]{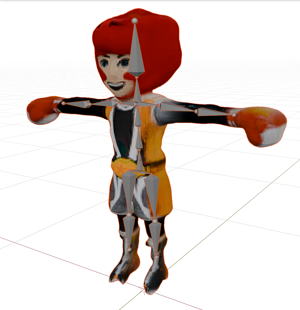}
			\centerline{10}
		\end{minipage}
		\begin{minipage}{0.09\linewidth}
			\centering
			\includegraphics[width=0.9\linewidth]{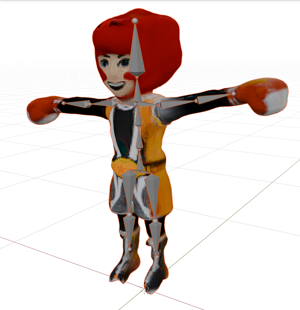}
			\centerline{20}
		\end{minipage}
		\begin{minipage}{0.09\linewidth}
			\centering
			\includegraphics[width=0.9\linewidth]{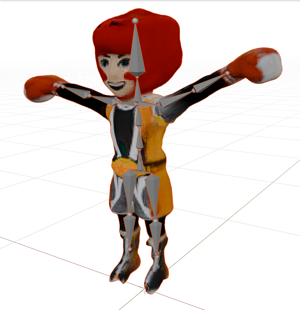}
			\centerline{30}
		\end{minipage}
		\begin{minipage}{0.09\linewidth}
			\centering
			\includegraphics[width=0.9\linewidth]{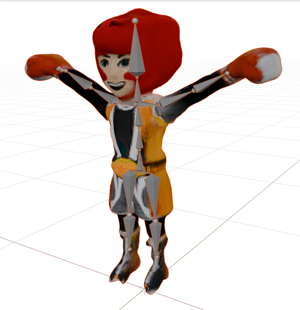}
			\centerline{40}
		\end{minipage}
		\begin{minipage}{0.09\linewidth}
			\centering
			\includegraphics[width=0.9\linewidth]{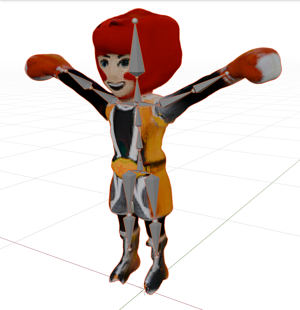}
			\centerline{50}
		\end{minipage}
		\begin{minipage}{0.09\linewidth}
			\centering
			\includegraphics[width=0.9\linewidth]{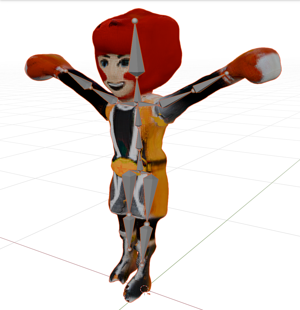}
			\centerline{60}
		\end{minipage}
		\begin{minipage}{0.09\linewidth}
			\centering
			\includegraphics[width=0.9\linewidth]{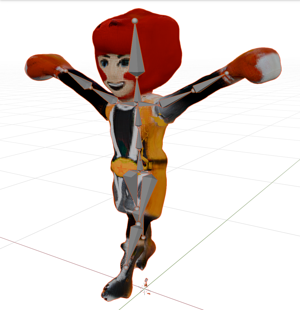}
			\centerline{70}
		\end{minipage}
		\begin{minipage}{0.09\linewidth}
			\centering
			\includegraphics[width=0.9\linewidth]{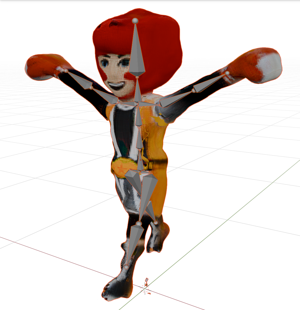}
			\centerline{80}
		\end{minipage}
		\qquad
		
		\caption{Multiple models self-adaption and generation evaluation. The first column represents the single input image, the other columns with number index represent the key-frame actions. Each row represents one same generated 3D model except first column.}
		\label{fig13}
	\end{figure}

	Generalization ability evaluation is an important item for the test of novel principle or method. To test the robust processing performance breadth and adaptability, as shown in Figure\ref{fig13}, we used enough numbers of single image to test the self-adaption armature generation and binding of OneTo3D, from which selecting 9 representative cases for the analyzing and evaluation. For each test 3D model, we used the action command text as same as shown in Figure\ref{fig12}. As shown in Figure\ref{fig13}, with the same specific user input command text, the 3D model will generate the same corresponding action in the same specific key-frame, which means the 3D model actions and 3D video generation methods of OneTo3D are model-independent. The key-frames inserting regulation is same as shown in Figure\ref{fig12}. Therefore, the 3D model actions and 3D video generation methods of OneTo3D have robust performance breadth and adaptability. 
	
	As shown in Figure\ref{fig13}, the models with different shapes will have different self-adaption armature generating and binding performances. One advantage of using 3D model armature to control and generate 3D model actions is repeatability of a series of complex actions. Meanwhile, the self-adaption generation and binding of the armature promotes the re-editable design of 3D model actions and 3D video generation. Generally, the input 3D model image more complex and detailed, the generation and  motion performance of the self-adaption armature will be more difficult.

\subsection{Fast 3D Video Generation}
	
To test the 3D video generating speed of OneTo3D, we used the models shown in Figure\ref{fig13} to test the 3D video generating speed with the same command. The models in Table\ref{tb1} are named as from $model1$ to $model9$ by rows in Figure\ref{fig13}. As shown in Table\ref{tb1}, each experiment will generate a 3D video with $80$ frames that includes $8$ key-frames (represented by $KFs$). Parameters $RT_i, i \in [1, 2, 3, 4, 5]$ represent 5 times experiment results of evaluating the 3D video generating spent time in seconds. $avgRT$ means the average generating time of 3D video in seconds. $speed F/s$ means the numbers of generated frames per second. $\tau_{x}, \tau_{y}, \tau_{z}$ is the localization fine-tuning factors in direction $x, y, z$ from Equation\ref{eq7}.

As shown in Table\ref{tb1}, for a computer or machine that has fixed calculating resource, $avgRT$ is approximate. The average of the $speed F/s$ is 8.8086 frames per second in generating. By testing and analyzing the values of $\tau_{x}$, $\tau_{y}$ and $\tau_{z}$, we found that for most of the input images that have a simple shape or structure support the default fine-tuning factor $(\tau_{x}, \tau_{y}, \tau_{z}) = (0, 0, 0)$. For the input object images that have a non-regular trunks proportion, to avoid the mismatching or invalidation of the self-adaption armature binding, fine-tuning the value of $\tau_{z}$ will make issue feasible in most cases. In the column of $\tau_{z}$, the bold figure means the value used in the experiment show in the current row. Other numbers are optional and feasible. $\tau_{z}) = 0$ will not always useful (e.g.: $\tau_{z}$ in model8) but is feasible in most cases.

\begin{table}[H]
	\centering
	\caption{3D Video Generating Speed with Command Actions}
	\begin{tabular}{ccccccccccccc}
		\toprule
		name & frames & KFs & RT1/s & RT2/s & RT3/s & RT4/s & RT5/s & avgRT/s & speed F/s  & $\tau_{x}$  &  $\tau_{y}$  &  $\tau_{z}$   \\
		\midrule
		model1	& 80  &  8  &  9.85  &	 9.04  &  8.95  &  8.96  &	 8.92  &  9.144   &  8.7489  &  0	&  0  &   0 /\textbf{0.05}   \\
		model2	& 80  &  8  &  9.03  &	 9.00  &  9.07  &  9.03  &	 9.49  &  9.124   &  8.7681  &  0	&  0  &   0 /\textbf{0.05}   \\
		model3  & 80  &  8  &  9.89  &	 8.91  &  9.05  &  9.47  &	 9.05  &  9.274   &  8.6263  &  0	&  0  &   0/\textbf{0.05}   \\
		model4  & 80  &  8  &  9.00  &	 8.96  &  8.97  &  9.68  &	 8.97  &  9.116   &  8.7758  &  0	&  0  &   \textbf{0.10}   \\
		model5  & 80  &  8  &  9.08  &	 9.82  &  8.95  &  8.97  &	 8.95  &  9.154   &  8.7393  &  0	&  0  &   \textbf{0.05}   \\
		model6  & 80  &  8  &  8.97  &	 9.05  &  9.01  &  8.90  &	 8.91  &  8.968   &  8.9206  &  0	&  0  &   0/\textbf{0.05}   \\
		model7  & 80  &  8  &  9.22  &	 9.01  &  8.95  &  9.05  &	 8.94  &  9.034   &  8.8554  &  0	&  0  &   0.05/\textbf{0.10}   \\
		model8  & 80  &  8  &  9.08  &	 8.95  &  9.07  &  8.88  &	 8.99  &  8.994   &  8.8948  &  0	&  0  &   \textbf{0.27}/0.30   \\
		model9  & 80  &  8  &  8.96  &	 9.01  &  8.87  &  9.04  &	 8.82  &  8.940   &  8.9485  &  0	&  0  &   0/\textbf{0.05}    \\
		\bottomrule
	\end{tabular}
	\label{tb1}    
\end{table}

\subsection{Generation Characteristics Comparison}

To better evaluate the advantages of OnTo3D, we compared OnTo3D with the new SOTA projects in generation. As shown in Table\ref{tb2}, $C-SD$ represents conditioned diffusion model. $Dyn$ means the dynamic. $Edit$ means editable. $M$ represents being able to generate 3D model. $V$ presents being able to generate 3D video. The $Speed$ is rendering and generating speed for 3D model or 3D video. $Resol$ means the generation resolution. $CD$ and $3D IoU$ (Intersection Over Union) are comparison of 3D metrics. $mVRAM$ represents the influencing or training VRAM(Video RAM) minimum required usage of GPU. $\surd$ means possessing, $\otimes$ means lacking. $\surd+$ means possessing and better. In this $Speed$ column, $m$ represents the minutes, $s$ represents the seconds, $Ss+$ means less than a minute. $+$ means more, $-$ means less, and $\pm$ means floating up or down. In this column $Base$, displaying the basic implementation methods of the compared projects. $3D-SD$ means 3D Stable Diffusion. $2D-SD$ means 2D Stable Diffusion. $Train$ means using the machine learning methods but not using the $SD$ models. $Con$ means developing based on controlling armature.

\begin{table}[H]
	\centering
	\caption{Generation Characteristics Comparisons with Other SOTA Models}
	\begin{tabular}{ccccccccccccc}
		\toprule
		Model & Dyn & Edit &  M  &  V  &  Speed  &  Image  &  Base  & Resol  & CD $\downarrow $ & 3D IoU $\uparrow$  & mVRAM  \\
		\midrule
		Sv3d\cite{voleti2024sv3d}  & $\otimes$  &  $\surd$   &  $\surd$   &  $\surd$   &  2-8m+  &  one  &  3D-SD  &  ****  &  0.024  &  0.614  &  8x80GB  \\-
		
		Point-E \cite{nichol2022point}  & $\otimes$  &  $\surd$   &  $\surd$   &  $\otimes$   &  1-2m  &  syn   &  SD  &  ***  & 0.074 &  0.162  & 1 GPU   \\
		
		Shap-E \cite{jun2023shap}  & $\otimes$  &  $\surd$   &  $\surd$   &  $\otimes$   &  Ss$\pm$  &  syn  &  SD  &  ***+  &  0.071  &  0.267  &  1 V100   \\
		
		DreamGaussian \cite{tang2023dreamgaussian}  & $\otimes$  &  $\surd$   &  $\surd$   &  $\otimes$   &  -2m  &  one  &  GS  &  ***  &  0.055  &  0.411  &  8-16GB+  \\
		
		One-2-3-45++  \cite{voleti2024sv3d}  & $\otimes$  &  $\surd$   &  $\surd$   &  $\otimes$   &  1m$\pm$  &  one  &  2D-SD  &  ****   &   0.054  &  0.406  &  8xA100  \\
		
		EscherNet  \cite{kong2024eschernet}  & $\otimes$  &  $\surd$   &  $\surd$   &  $\otimes$   &  $**$  &  one+  &  C-SD  &  ****  &   0.042  &  0.466  &  6xA100   \\
		
		Free3D  \cite{zheng2023free3d}  & $\otimes$  &  $\surd$   &  $\surd$   &  $\otimes$   &  Ss$\pm$  &  syn  &  Train  &  ****   &  0.053  &  0.451  &   4×A40:48  \\
		
		Stable Zero123  \cite{liu2023zero}  & $\otimes$  &  $\surd$   &  $\surd$   &  $\otimes$   &  Ss$\pm$   &  one  &  SD  &  ****  &  0.047  &  0.426  &  6xA100     \\
		
		\midrule
		OneTo3D   & $\surd$  &  $\surd+$   &  $\surd$   &  $\surd$   &  Ss$\pm$  &  one  &  Con  &   ***  &  0.050$\pm$  &  0.910$\pm$  &   0-8GB+   \\
	
		\bottomrule
	\end{tabular}
	\label{tb2}   
\end{table}

As shown in Table\ref{tb2}, OneTo3D has its advantages in implementing the dynamic, re-editable generation of 3D models and 3D videos. Compared with the existed previous SOTA Sv3d\cite{voleti2024sv3d}, OnTo3D has faster generating speed and more precise control, meanwhile, requires less VRAM and some of better 3D comparison metric performances. Therefore, for the normal basic 3D models and videos generation, the idea of OneTo3D is worth consideration. Compared with other projects, in terms of functional richness and real-time usability, OneTo3D is more advanced and practical.

\section{Conclusion}

In this paper, we state and analyze the novel design idea of OneTo3D. After an abundant but relative background introduction and background discussion, we make a detailed interpretation of the implementation mechanism of OneTo3D. With enough demo and open source code, we believe that the novel idea of OneTo3D is useful and effective. Compared with the pure implicit generation method, OneTo3D is more easier in precise controlling. OneTo3D has three main advantages: the first is OneTo3D can implement a dynamic 3D model generation; the second is Oneto3D can implement the re-editable more precise control operations; the three is OneTo3D can generate a unlimited time semantic continue 3D video with special text commands interpreter mechanism. Therefore it will be a good attempt to try the OneTo3D. Further optimization algorithms and design will be presented in next version of OneTo3D.

	
	\bibliographystyle{ACM-Reference-Format}
	\bibliography{sample-base}

	
	
	
	
	
	
	
	
	
\end{document}